\documentclass[review,number]{elsarticle}
\biboptions{sort&compress}

\usepackage{amssymb}
\usepackage{amsmath}

\journal{Aerospace Science and Technology}

\usepackage{amsmath,amssymb}
\usepackage{graphicx}
\usepackage{booktabs}
\usepackage{bm}

\journal{Aerospace Science and Technology}

\begin{document}

\begin{frontmatter}

\title{Saturation-Aware Robust Trajectory Optimization for Reusable Launch Vehicles via Differentiable Physics}

\author[1]{Liwei Chen\corref{cor1}}
\ead{liwei.chen.aero@gmail.com}

\author[1]{Tong Qin}

\cortext[cor1]{Corresponding author}

\address[1]{Beijing Institute of Aeronautical Systems Engineering, Beijing 100076, China}
\begin{abstract}
The high-angle-of-attack flip maneuver of reusable launch vehicles presents significant challenges for robust trajectory optimization due to the combined effects of highly nonlinear dynamics, aerodynamic uncertainties, and actuator saturation. This paper presents a differentiable physics framework for saturation-aware robust trajectory optimization. At its core, a Differentiable Particle Tube Control (DPTC) scheme is developed to optimize uncertainty evolution through an ensemble-based distribution shaping strategy. State uncertainty is represented by a Lagrangian particle ensemble, while hard actuator projection operators are embedded directly into the computational graph, enabling the joint optimization of the nominal feedforward trajectory and a time-varying feedback policy via end-to-end backpropagation. The proposed framework is evaluated against an automatic differentiation-based Successive Convexification (AD-SCvx) baseline combined with a conventional covariance steering feedback strategy. Six-degree-of-freedom Monte Carlo simulations demonstrate that, although the baseline achieves nominal fuel-optimal solutions, its unconstrained feedback formulation becomes susceptible to actuator saturation under aerodynamic disturbances, leading to degraded closed-loop robustness. In contrast, the proposed DPTC framework proactively performs a constraint-aware performance trade-off by relaxing spatial tracking to preserve critical control authority. Consequently, it prevents actuator saturation-induced performance degradation, reduces the 50\% Circular Error Probable (CEP$_{50}$) by 87\%, and significantly improves terminal landing precision while maintaining strict physical constraint satisfaction. These results demonstrate that integrating differentiable physics with ensemble-based optimization provides an effective and practical framework for robust guidance in highly constrained aerospace flight systems.
\end{abstract}


\end{frontmatter}

\section{Introduction}
\label{sec:introduction}
The advent of reusable launch vehicles (RLVs) has fundamentally transformed the aerospace industry, driving a paradigm shift toward economical and sustainable space access. For next-generation super-heavy reusable launch vehicles (RLVs)---exemplified by the Starship architecture---the vertical powered landing (VPL) phase represents one of the most critical and dynamically challenging segments of the flight profile~\cite{blackmore2016, lu2018}. Unlike traditional slender rockets, these vehicles perform a high-angle-of-attack ``belly-flop'' atmospheric entry to dissipate kinetic energy, followed by an aggressive and highly nonlinear ``flip maneuver'' to achieve a vertical orientation prior to touchdown. During this transient phase, the vehicle is subjected to intense aerodynamic coupling, large structural loads, and rapid variations in dynamic pressure~\cite{steinfeldt2010}. Thus, the trajectory becomes highly sensitive to environmental perturbations, including unsteady aerodynamic coefficients, wind shear, and mass-property uncertainties. Ensuring extreme landing precision and vehicle survivability under such severe multi-source uncertainties remains an open challenge in modern aerospace guidance, navigation, and control (GNC)~\cite{malyuta2022}.

To address the stringent requirements of VPL, Successive Convexification (SCvx) has emerged as a cornerstone methodology, demonstrating remarkable success by iteratively approximating non-convex dynamics into tractable convex subproblems~\cite{acikmese2007, mao2018, szmuk2020}. To account for exogenous disturbances, robust variants such as Tube-based Model Predictive Control (Tube-MPC) and Covariance Steering (CS) have been integrated. These methods construct a safety tube around a nominal trajectory and employ Riccati-based feedback laws to suppress deviations~\cite{mayne2005, okamoto2019, ridderhof2020}. 

Despite their theoretical appeal, classical robust SCvx frameworks face two fundamental limitations in extreme maneuvers. First, they rely heavily on local linearization and Gaussian covariance propagation~\cite{mesbah2016}, which may overly compress the true non-Gaussian state distributions induced by severe nonlinear aerodynamic transients, rendering the approximation insufficient for critical tail events~\cite{korda2018}. Second, they suffer from \textit{saturation blindness}. When a severe disturbance drives the required feedback command beyond the strict mechanical bounds (e.g., maximum thrust or TVC gimbal limits), 
actuator saturation occurs, instantly violating the theoretical robustness guarantees~\cite{astrom1989windup, tarbouriech2011saturation, nguyen2021pio}.

Parallel to these developments, advances in deep learning have introduced Differentiable Predictive Control (DPC) and Neural Ordinary Differential Equations (Neural ODEs), enabling exact gradient computation through complex dynamic graphs without analytical Jacobians. In our preliminary work~\cite{chen2025diff}, we established a differentiable Neural ODE framework that tightly coupled a neural aerodynamic surrogate with rigid-body dynamics, achieving high-fidelity fuel-optimal flip maneuvers. However, similar to most existing DPC paradigms, that framework was confined to deterministic nominal trajectory optimization. Even when disturbances are considered, current DPC methods typically rely on pointwise state penalties or low-order summary statistics (mean and covariance), rather than governing the evolution of the entire uncertainty distribution. 

A critical gap remains: current methodologies act as trajectory optimizers that attempt to rigidly track a nominal path. In contrast, under severe uncertainties and strict actuator limits, the optimal strategy should actively shape the stochastic state distribution through nonlinear dynamics.

Motivated by the concept of probability transport in generative modeling~\cite{lipman2023, albergo2023}, this paper proposes a novel framework termed \textbf{Differentiable Particle Tube Control (DPTC)}. Rather than treating robust trajectory optimization as a deterministic boundary-value problem, DPTC formulates it as a constrained stochastic probability transport problem under strict physical actuator limits. By representing the uncertain flight dynamics via a Lagrangian particle ensemble, DPTC learns a saturation-aware feedback policy that actively transports the initial non-Gaussian state distribution toward a compact, safe terminal measure. 

The main contributions of this paper are summarized as follows:
\begin{enumerate}
    \item \textbf{Transition from Trajectory Optimization to Distribution Shaping:} Rather than relying on classical Gaussian-linearization approximations, we introduce a framework that operates directly on the exact nonlinear dynamics. By leveraging a natively differentiable 6-DoF physics engine, DPTC shapes the stochastic probability manifold, inherently capturing rare-event dispersions and non-Gaussian aerodynamic responses.
    \item \textbf{Native Saturation-Aware Feedback Synthesis:} Unlike covariance steering which appends feedback post-hoc without saturation awareness, DPTC embeds hard physical actuator limits (e.g., $\text{Proj}_{\mathcal{U}}$) directly into the forward computational graph. The optimizer simultaneously synthesizes the feedforward sequence ($\boldsymbol{U}_{\text{ref}}$) and the feedback policy ($\mathcal{K}$) while ``experiencing'' the same actuator saturations that govern the real vehicle.
    \item \textbf{Soft-Penalty Tail Risk Regulation:} To enforce operational safety without the brittleness of hard chance-constraints, we propose a differentiable tail-risk objective. This ensures smooth gradient landscapes for backpropagation through time (BPTT) while aggressively penalizing constraint-violating particles, yielding a highly survivable closed-loop guidance law.
\end{enumerate}

The remainder of this paper is organized as follows. Section~\ref{sec:problem_formulation} formulates the 6-DoF flip-landing dynamics and the unified differentiable physics engine. Section~\ref{sec:methodology} explicitly contrasts the algorithmic architectures of AD-SCvx / covariance steering AD-SCvx (CS-AD-SCvx)  and the proposed DPTC. Section~\ref{subsec:nominal_results} presents the deterministic optimization limits, while Section~\ref{sec:results} validates the robust probability transport performance under severe aerodynamic uncertainties through closed-loop Monte Carlo simulations. Finally, Section~\ref{sec:conclusions} concludes the paper.

\section{Problem Formulation and the Differentiable Physics Engine}
\label{sec:problem_formulation}

The foundational pillar of both optimization paradigms evaluated in this study is the native Differentiable Physics Engine. It intrinsically couples the 6-DoF rigid-body kinematics in the $SO(3)$ manifold with a deep-learned unsteady aerodynamic surrogate. By formulating all physical state transitions as differentiable tensor operations, this engine seamlessly bridges rigorous aerospace kinematics with modern autograd-based deep learning frameworks.

\subsection{6-DoF Flight Dynamics}
\label{subsec:6dof_dynamics}
Let the vehicle state vector be denoted as $\boldsymbol{x} = [\boldsymbol{r}^T, \boldsymbol{v}^T, \boldsymbol{q}^T, \boldsymbol{\omega}^T, m, \delta_y, \delta_z]^T \in \mathbb{R}^{16}$, where $\delta_y$ and $\delta_z$ explicitly represent the Thrust Vector Control (TVC) gimbal angles defined relative to the vehicle's body-fixed frame. The exact physical control input is defined as $\boldsymbol{u} = [T_{\text{cmd}}, u_{\delta y}, u_{\delta z}]^T \in \mathbb{R}^3$, comprising the thrust magnitude command and the respective gimbal angle rate commands. The continuous-time flight dynamics is governed by coupled ordinary differential equations (ODEs).
\begin{equation}
    \begin{cases}
        \dot{\boldsymbol{r}} = \boldsymbol{v} \\
        \dot{\boldsymbol{v}} = \frac{1}{m} (\boldsymbol{F}_{\text{aero}} + \boldsymbol{F}_{\text{thrust}}) + \boldsymbol{g} \\
        \dot{\boldsymbol{q}} = \frac{1}{2} \boldsymbol{q} \otimes [0, \boldsymbol{\omega}^T]^T \\
        \dot{\boldsymbol{\omega}} = \boldsymbol{J}^{-1} (\boldsymbol{M}_{\text{aero}} + \boldsymbol{M}_{\text{thrust}} - \boldsymbol{\omega} \times \boldsymbol{J}\boldsymbol{\omega}) \\
        \dot{m} = -\frac{T_{\text{cmd}}}{I_{\text{sp}} g_0}, \quad \dot{\delta}_y = u_{\delta y}, \quad \dot{\delta}_z = u_{\delta z}
    \end{cases}
\end{equation}
where $\boldsymbol{J}$ is the inertia tensor, $\boldsymbol{q}$ is the attitude quaternion, and $\boldsymbol{F}$ and $\boldsymbol{M}$ denote the corresponding force and moment vectors.

\textbf{Remark on Control Space Dimensionality:} 
It is worth noting that the proposed DPTC framework operates strictly within this exact 3-dimensional physical control space ($\boldsymbol{u} \in \mathbb{R}^3$). In contrast, to accommodate the rigid requirements of the interior-point convex solvers, the baseline CS-AD-SCvx architecture must introduce an auxiliary slack variable $\sigma$ (where $\sigma \ge T_{\text{cmd}}$) to perform lossless convexification in the mass depletion dynamics (i.e., $\dot{m} = -\frac{\sigma}{I_{\text{sp}} g_0}$). This artificially lifts the CS-AD-SCvx control space to $\mathbb{R}^4$, a mathematical workaround entirely bypassed by the native nonlinear autograd engine in DPTC.

\textbf{Propulsion System and Engine Switching Logic:}
The propulsion system model accounts for the discrete staging logic of the multi-engine configuration. To ensure numerical continuity during gradient-based optimization, the number of active engines $n_e$ is modeled as a function of the vehicle altitude $y$. We define the transition thresholds as $h_1 = 387$ m and $h_2 = 487$ m. The engine count is governed by the following continuous mapping:

\begin{equation}
n_e(y) = 
\begin{cases} 
2.0, & y < h_1 \\
2.0 + \frac{y - h_1}{h_2 - h_1}, & h_1 \le y \le h_2 \\
3.0, & y > h_2
\end{cases}
\end{equation}

where the linear transition within $[h_1, h_2]$ serves as a differentiable surrogate for the discrete switching event. This formulation effectively avoids the gradient discontinuities inherent in traditional hard-switching logic, ensuring a smooth loss landscape suitable for Backpropagation Through Time (BPTT). Hence, the total thrust magnitude $T$ is computed as $T = n_e(y) \cdot T_{\text{unit}}$, where $T_{\text{unit}}$ is the commandable thrust of a single engine unit.

\subsection{Neural Aerodynamic Surrogate and Scope of Fidelity}
\label{subsec:aero_surrogate}
To capture the nonlinear dynamic stall during the extreme flip maneuver, a Multi-Layer Perceptron (MLP) surrogate is constructed to predict the aerodynamic force coefficients $\boldsymbol{C} = [C_l, C_d, C_m]^T$. 

To strictly isolate the performance evaluation of the proposed stochastic control methodologies from the extreme computational complexities of high-fidelity computational fluid dynamics (CFD), two deliberate simplifying assumptions are made for the aerodynamic surrogate:
\begin{itemize}
    \item \textbf{Axisymmetric Equivalent Formulation:} The aerodynamic map assumes an axisymmetric vehicle mold line. When lateral velocity components (sideslip) emerge, the 3D velocity vector is projected to compute an equivalent spatial angle of attack (AoA), $\alpha_{\text{eq}} = \arccos(\frac{\boldsymbol{v}^T \boldsymbol{x}_{\text{body}}}{\|\boldsymbol{v}\|})$.
    \item \textbf{Quasi-Steady Assumption:} The current surrogate focuses solely on the static nonlinearities across the Mach-AoA envelope. The integration of dynamic pitch damping derivatives (e.g., dependence on the non-dimensional pitch rate $\hat{q}$) is reserved for future high-fidelity iterations.
\end{itemize}
Then the aerodynamic mapping is simplified to:
\begin{equation}
    \boldsymbol{C} = \text{MLP}(\cos\alpha_{\text{eq}}, \sin\alpha_{\text{eq}}; \boldsymbol{\Theta})
\end{equation}
Detailed discussions regarding the CFD data generation, MLP architecture, and offline training procedures are provided in Appendix A and our preliminary work \cite{chen2025diff}.

\subsection{Differentiable Integration and Stochastic Injection}
\label{subsec:diff_integration}
The neural surrogate is deeply embedded into the continuous-time stochastic dynamics $\dot{\boldsymbol{x}} = \boldsymbol{f}(\boldsymbol{x}, \boldsymbol{u}, \boldsymbol{w})$, where $\boldsymbol{w}$ denotes exogenous non-steady aerodynamic perturbations (e.g., wind gusts and parametric uncertainties). 

To enable end-to-end gradient propagation for both offline learning and covariance propagation, the system is discretized using a fixed-step 4th-order Runge-Kutta (RK4) integrator:
\begin{equation}
    \boldsymbol{x}_{k+1} = \boldsymbol{F}_{RK4}(\boldsymbol{x}_k, \boldsymbol{u}_k, \boldsymbol{w}_k, \Delta t)
\end{equation}
Because all mathematical operations—including the MLP forward pass, quaternion normalizations, and rigid-body ODEs within $\boldsymbol{F}_{RK4}$—are formulated exclusively using differentiable tensor operations via PyTorch, the entire physics engine acts as a native autograd computational graph. This structural differentiability forms the critical prerequisite for both the Covariance Steering Jacobian extraction (Method I) and the DPTC probability transport backpropagation (Method II).

\subsection{Mission Profile and Boundary Conditions}
\label{subsec:mission_profile}

To systematically evaluate the efficacy of the trajectory optimization paradigms, a challenging 6-DoF Starship-class flip maneuver is investigated. The vehicle is initialized in a high-altitude, windward-glide configuration representing the ``belly-flop'' phase. The initial state is defined at an altitude of 750 m, with a substantial sink rate ($v_{x,0} = -18.82$ m/s, $v_{y,0} = -106.73$ m/s) and a near-horizontal pitch attitude of $170^\circ$. 

The target landing state requires the vehicle to reach the origin pad ($x_f=0, z_f=0, y_f=0$) with a strictly vertical attitude ($90^\circ$) and a soft-touchdown vertical velocity of $-0.1$ m/s. Throughout the maneuver, the operational boundaries strictly enforce a maximum Thrust Vector Control (TVC) gimbal limit of $\pm 10^\circ$. This highly constrained mission profile establishes a rigorous baseline for assessing the fundamental trade-offs between mathematical optimality and engineering survivability.

\section{Methodology}
\label{sec:methodology}

Built upon the differentiable physics engine established in Section \ref{sec:problem_formulation}, this section details the algorithmic frameworks for robust trajectory optimization. Two fundamentally different mathematical architectures for uncertainty handling are constructed and evaluated: (i) local uncertainty regulation via automatic differentiation-based Gaussian covariance steering within a Successive Convexification architecture (CS-AD-SCvx), and (ii) direct stochastic distribution shaping through the proposed Differentiable Particle Tube Control (DPTC). 

While both leverage the shared automatic differentiation infrastructure to extract gradients through the nonlinear rigid-body dynamics and aerodynamic surrogates, they diverge significantly in their theoretical treatment of probability evolution and physical actuator saturation.

\subsection{Method I: AD-based Covariance Steering Successive Convexification (CS-AD-SCvx)}
\label{subsec:ad_scvx}

Sequential Convex Programming (SCvx) represents the industrial standard for deterministic trajectory optimization. To account for aerodynamic uncertainties within this classical framework, an Automatic Differentiation-based Covariance Steering architecture (CS-AD-SCvx) is established as the baseline. 

\vspace{0.5em}
\textbf{1. Local Linearization and Chance-Constrained Optimization}

This methodology approximates the evolution of the stochastic state distribution by enforcing a local Gaussian approximation around a deterministic nominal trajectory. Automatic Differentiation (AD) via mixed forward- and reverse-mode is utilized to extract the exact Jacobians of the nonlinear flight dynamics:

\begin{equation}
    \boldsymbol{A}_k = \left. \frac{\partial \boldsymbol{F}_{RK4}}{\partial \boldsymbol{x}} \right|_{\boldsymbol{x}_{k,\text{ref}}}, \quad
    \boldsymbol{B}_k = \left. \frac{\partial \boldsymbol{F}_{RK4}}{\partial \boldsymbol{u}} \right|_{\boldsymbol{u}_{k,\text{ref}}}, \quad
    \boldsymbol{G}_k = \left. \frac{\partial \boldsymbol{F}_{RK4}}{\partial \boldsymbol{w}} \right|_{\boldsymbol{w}_{k} = \boldsymbol{0}}
\end{equation}

To regulate the probability dispersion, a Linear Covariance (LinCov) steering mechanism is established. Under the Separation Principle, a sequence of unconstrained Linear Quadratic Regulator (LQR) feedback gains $\boldsymbol{K}_k$ is computed offline via the backward Riccati equation along the nominal trajectory. The closed-loop state transition matrix is defined as $\boldsymbol{A}_{\text{cl},k} = \boldsymbol{A}_k - \boldsymbol{B}_k \boldsymbol{K}_k$, and the dispersion covariance $\boldsymbol{\Sigma}_k$ is propagated forward iteratively:

\begin{equation}
    \boldsymbol{\Sigma}_{k+1} = \boldsymbol{A}_{\text{cl},k} \boldsymbol{\Sigma}_k \boldsymbol{A}_{\text{cl},k}^T + \boldsymbol{G}_k \boldsymbol{W} \boldsymbol{G}_k^T
\end{equation}

where $\boldsymbol{W}$ is the covariance of the aerodynamic noise. Utilizing this analytically approximated $\boldsymbol{\Sigma}_k$, the deterministic state constraints are tightened into probabilistic chance constraints. For instance, the safety pitch corridor is dynamically restricted by a $\kappa$-sigma margin:

\begin{equation}
    \theta_{\text{ref},k} \ge \theta_{\text{min}} + \kappa \sqrt{\nabla_{\boldsymbol{x}}\theta_k \, \boldsymbol{\Sigma}_k \, \nabla_{\boldsymbol{x}}\theta_k^T}
\end{equation}

The resulting Second-Order Cone Program (SOCP) balances fuel optimality and chance-constrained safety bounds under successive trust-region iterations.

\vspace{0.5em}
\textbf{2. Inherent Limitations under Extreme Transients} 

While computationally efficient and mathematically mature, CS-AD-SCvx encounters two theoretical limitations when applied to the extreme flip maneuver. 

First, the \textit{Local Linearization Assumption}: The Jacobian-based propagation dictates that disturbances propagate linearly. Under severe aerodynamic nonlinearities, this local Gaussian assumption may inaccurately capture the skewed evolution of true probability density functions, potentially generating overly conservative or artificially infeasible constraint tubes.

Second, \textit{Unconstrained Control Allocation}: Driven by the traditional separation principle, the Riccati feedback $\boldsymbol{K}_k$ is synthesized under the assumption of infinite control authority ($\boldsymbol{u} \in \mathbb{R}^m$). Although chance constraints proactively tighten the nominal state trajectory, the algorithm remains structurally unaware of physical actuator hard limits during feedback synthesis. When an exogenous gust demands a corrective effort ($\delta \boldsymbol{u}_k = -\boldsymbol{K}_k \delta \boldsymbol{x}_k$) that exceeds the engine's deep-throttling limit ($T_{\min}$) or TVC bounds, the physically executed commands are truncated. This hardware truncation invalidates the theoretical closed-loop covariance propagation, potentially precipitating an under-actuated stall.

\subsection{Method II: Differentiable Particle Tube Control (DPTC)}
\label{subsec:method_dptc}

To address the limitations of localized Gaussian approximation and unconstrained control allocation, the DPTC framework is proposed. DPTC formulates robust landing guidance as a constrained stochastic distribution shaping problem, directly learning a saturation-aware feedback policy by evaluating the evolution of the stochastic state distribution through the nonlinear flight dynamics.

\vspace{0.5em}
\textbf{1. Stochastic Distribution Shaping Formulation}

The uncertain flight dynamics are modeled as an It\^{o} stochastic differential equation (SDE),

\begin{equation}
d\mathbf{x}_t = \mathbf{f} (\mathbf{x}_t,\mathbf{u}_t) dt + \mathbf{\Sigma}_w(\mathbf{x}_t)d\mathbf{W}_t
\end{equation}

where $\mathbf{W}_t$ denotes the standard Wiener process representing aerodynamic disturbances. Therefore, the vehicle state becomes a stochastic process, whose probability density $p(\mathbf{x},t)$ mathematically evolves according to the corresponding Fokker--Planck--Kolmogorov (FPK) equation.

Note that in the present study the distribution shaping mechanism is realized through the end-to-end differentiable Monte Carlo rollout of a particle ensemble. Following this Lagrangian perspective, the probability density is computationally approximated by an ensemble of $N_p$ stochastic realizations:

\begin{equation}
p(\mathbf{x},t) \approx \frac{1}{N_p} \sum_{i=1}^{N_p} \delta ( \mathbf{x} - \mathbf{x}^{(i)}_t )
\end{equation}

where each particle experiences an independent disturbance realization while sharing a unified control policy. This transforms the infinite-dimensional distribution shaping problem into a finite, differentiable ensemble optimization problem leveraging vectorized GPU operations. When $N_p=1$, the empirical probability density collapses to a Dirac measure, naturally reducing the framework to a deterministic trajectory optimizer.

\vspace{0.5em}
\textbf{2. Saturation-Aware Feedback Policy}

The control policy is parameterized as a time-varying affine feedback law. The learnable parameters consist of the nominal feedforward sequence $\mathbf{U}_{\mathrm{ref}}$, the terminal time $t_f$, and the feedback gain sequence $\mathcal{K} = \{ \mathbf{K}_0, \dots, \mathbf{K}_{N-1} \}$.

For the $i$-th particle, the physically executed actuator command is computed as:

\begin{equation}
\mathbf{u}^{(i)}_k = \Pi_{\mathcal U} \left( \mathbf{u}_{k,\mathrm{ref}} - \mathbf{K}_k ( \mathbf{x}^{(i)}_k - \mathbf{x}_{k,\mathrm{ref}} ) \right)
\end{equation}

where $\Pi_{\mathcal U}(\cdot)$ denotes the projection operator enforcing absolute engine thrust and gimbal limits. By embedding this hard truncation directly into the differentiable forward simulation, the optimizer simultaneously synthesizes $\mathbf{U}_{\mathrm{ref}}$ and $\mathcal{K}$ under strict physical actuation bounds, ensuring the resulting policy is saturation-aware.

\vspace{0.5em}
\textbf{3. Ensemble Rollout and Distribution Evolution}

The deterministic reference trajectory is propagated without disturbances. Simultaneously, every particle in the Lagrangian ensemble is propagated through the nonlinear flight dynamics:

\begin{equation}
\mathbf{x}^{(i)}_{k+1} = \mathbf{F}_{\mathrm{RK4}} \left( \mathbf{x}^{(i)}_k, \mathbf{u}^{(i)}_k, \mathbf{w}^{(i)}_k, \Delta t \right), \qquad i=1,\ldots,N_p
\end{equation}

This batch rollout provides a differentiable particle approximation of the controlled distribution evolution governed by the aerodynamics.

\vspace{0.5em}
\textbf{4. Terminal Constraints and Tail-Risk Optimization}

Landing accuracy and operational safety are formulated into a continuous differentiable optimization landscape. The concentration of the terminal probability distribution is quantified through the first two statistical moments:

\begin{equation}
\mathcal L_{\mathrm{terminal}} = \| \boldsymbol{\mu}_N - \mathbf{x}_{\mathrm{target}} \|^2_{\mathbf Q_1} + \mathrm{Tr} ( \mathbf Q_2 \boldsymbol{\Sigma}_N )
\end{equation}

where $\boldsymbol{\mu}_N$ and $\boldsymbol{\Sigma}_N$ are computed empirically from the terminal particle ensemble.

To enforce operational boundaries, a differentiable tail-risk functional is introduced:

\begin{equation}
\mathcal L_{\mathrm{risk}} = \frac1{N_p} \sum_{i=1}^{N_p} \sum_{k=0}^{N-1} \phi \left( g( \mathbf{x}^{(i)}_k, \mathbf{u}^{(i)}_k ) \right)
\end{equation}

where $g(\cdot)\le0$ represents generic safety constraints. In this work, $\phi(z) = \mathrm{ReLU}^2(z)$ is adopted as a smooth tail-risk surrogate. This selectively penalizes probability mass entering unsafe regions while preserving a continuous gradient landscape suitable for backpropagation.

The complete penalized optimization problem is defined as:

\begin{equation}
\min_{\mathbf U_{\mathrm{ref}}, \mathcal K, t_f} \; \mathcal L_{\mathrm{total}} = \mathcal L_{\mathrm{terminal}} + \lambda_r \mathcal L_{\mathrm{risk}} + \gamma \sum_{k=0}^{N-1} \| \mathbf K_k \|_F^2
\end{equation}

Algorithmic gradients with respect to $\mathbf U_{\mathrm{ref}}$, $\mathcal K$, and $t_f$ are obtained via Backpropagation Through Time (BPTT) across the differentiable ensemble rollout and optimized using the Adam algorithm. 
For full reproducibility, the exact hyperparameter configurations, including the tier-scaled risk weights $\lambda_r$, learning rate schedules, and baseline chance-constraint margins, are detailed in Appendix \ref{app:hyperparameters}.

\vspace{0.5em}
\textbf{5. Remark on Subgradient Propagation and Optimizability}

The integration of the hard actuator projection operator $\Pi_{\mathcal U}(\cdot)$ and the tail-risk surrogate $\phi(\cdot)$ introduces structural non-smoothness into the computational graph. While these functions are not globally continuously differentiable ($C^1$), they are Lipschitz continuous and thus differentiable almost everywhere according to Rademacher's theorem. During Backpropagation Through Time (BPTT), the autograd engine evaluates the generalized Clarke subdifferential for these non-smooth operations. 

From an optimization perspective, a critical challenge arises when a Lagrangian particle deeply violates the physical limits: the local weak derivative of the clamping operator strictly evaluates to zero ($\partial \Pi_{\mathcal U} / \partial \boldsymbol{u} = \boldsymbol{0}$). This naturally halts the gradient flow of the downstream terminal tracking loss ($\mathcal L_{\mathrm{terminal}}$), a phenomenon equivalent to actuator saturation rendering the vehicle locally uncontrollable. 

However, the proposed formulation preserves robust optimizability through the differentiable tail-risk functional $\mathcal L_{\mathrm{risk}}$. Because $\phi(z) = \mathrm{ReLU}^2(z)$ is mathematically $C^1$-continuous and monotonically increasing within the violation region, it bypasses the zero-gradient block of the physical projection operator. It directly backpropagates a non-zero restorative gradient to the policy parameters ($\mathbf{U}_{\mathrm{ref}}, \mathcal{K}$), forcing the optimizer to actively reshape the probability distribution away from the non-differentiable saturation boundaries before the control authority is entirely exhausted.

\section{Results and Discussion}
\label{sec:results}

To ensure a consistent comparative analysis, both the CS-AD-SCvx baseline and the DPTC framework operate on the identical PyTorch-based differentiable physics engine, sharing the same aerodynamic surrogate, mass properties, and operational boundaries. Algorithmically, CS-AD-SCvx utilizes the CLARABEL interior-point solver for linearized SOCP subproblems, whereas DPTC employs the Adam optimizer to perform gradient descent directly on the un-linearized simulation rollout.

\subsection{Nominal Trajectory Optimization: The Deterministic Limit and Algorithmic Validation}
\label{subsec:nominal_results}

Before evaluating robust guidance under uncertainties, the deterministic limits of both frameworks are validated ($\boldsymbol{W} \to \boldsymbol{0}$). Under zero-disturbance conditions, the feedback matrices ($\boldsymbol{K}$) are decoupled. DPTC naturally reduces to a continuous direct-shooting trajectory optimizer ($N_p = 1$), and CS-AD-SCvx reverts to a standard deterministic SCvx solver ($\boldsymbol{\Sigma}_k = \boldsymbol{0}$). This phase cross-validates the physical fidelity of the DPTC autograd engine against the established SCvx baseline and highlights the operational limitations of purely deterministic planning.

\begin{figure}[htbp]
    \centering
    \includegraphics[width=0.48\textwidth]{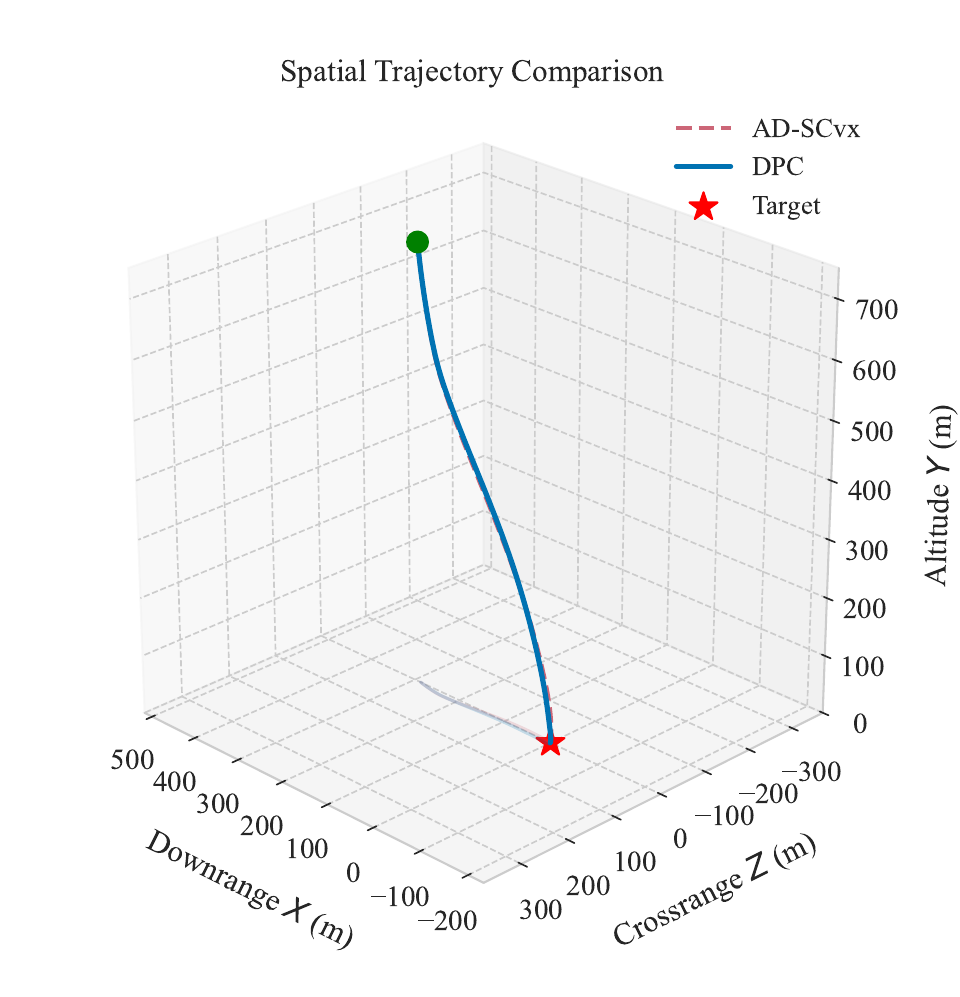}
    \includegraphics[width=0.48\textwidth]{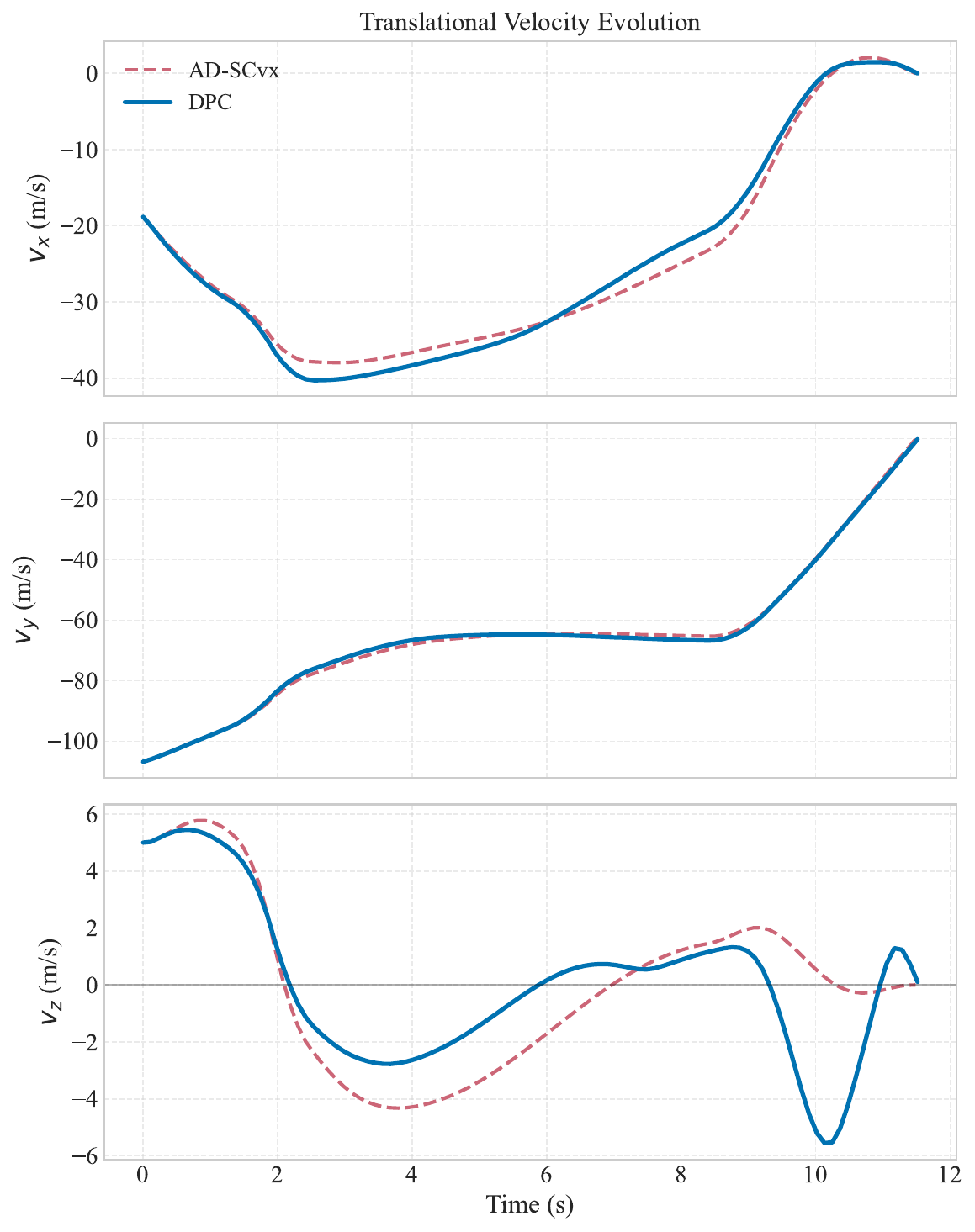}
    \caption{(Left) Comparative 3D spatial trajectories of the flip maneuver. (Right) Time history of the velocity components.}
    \label{fig:trajectory_and_velocity}
\end{figure}

\begin{figure}[htbp]
    \centering
    \includegraphics[width=0.7\textwidth]{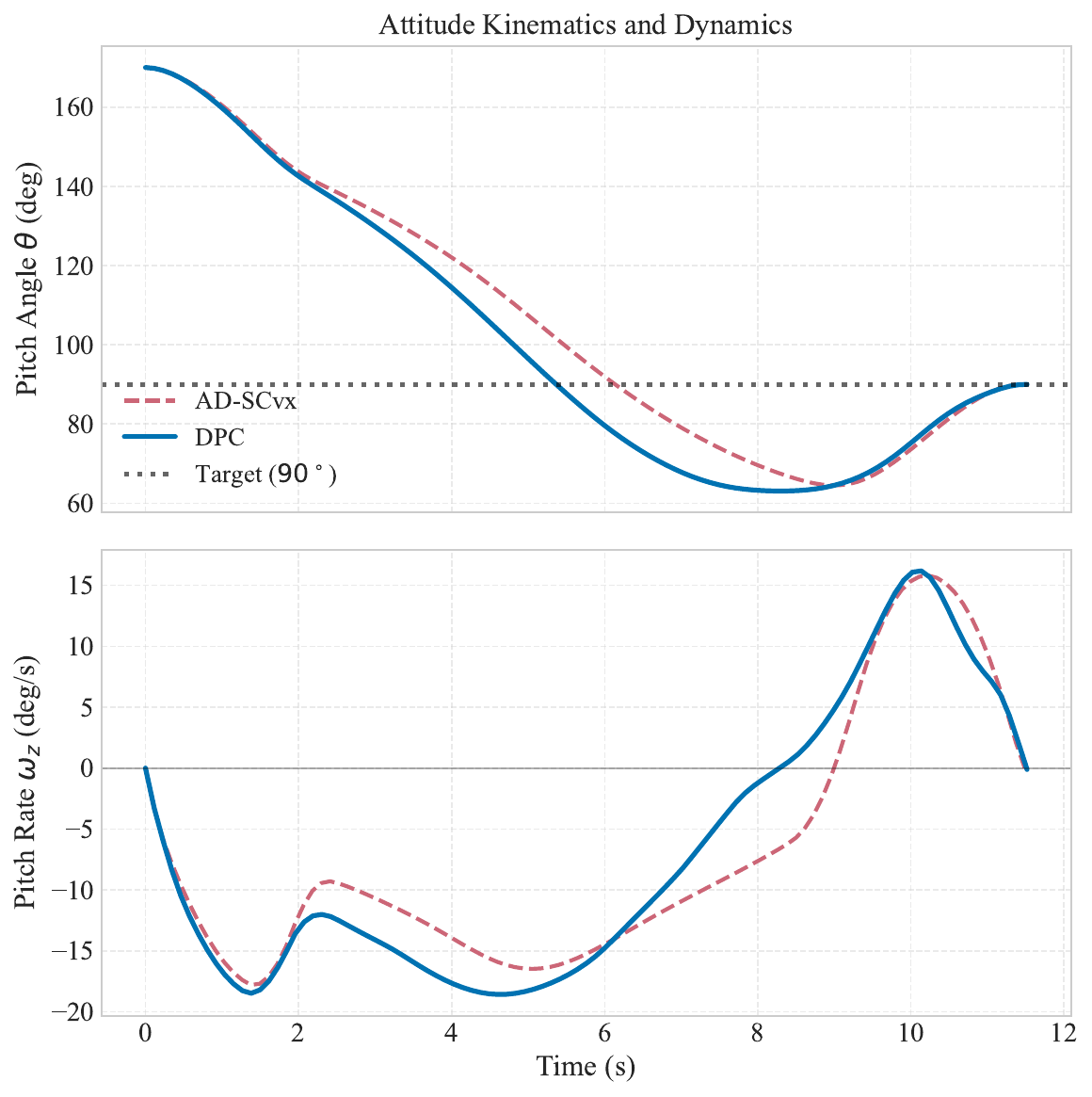}
    \caption{Evolution of the nominal pitch angle ($\theta$) and pitch rate ($\omega_z$), verifying the trajectory generation fidelity of the differentiable environment.}
    \label{fig:attitude_compare}
\end{figure}

\textbf{1. Kinematic Evolution and Baseline Matching} \\
As shown in \textbf{Fig.~\ref{fig:trajectory_and_velocity}}, both methods successfully guide the vehicle to the landing pad, transitioning smoothly from a high-speed glide to a vertical powered descent. The spatial trajectories exhibit high congruence. A minor crossrange (Z-axis) discrepancy emerges due to the relatively flat objective landscape along this axis, reflecting the distinct search mechanisms of the two optimizers (sequential trust-region vs. unconstrained gradient descent). Furthermore, the pitch and pitch rate trajectories (\textbf{Fig.~\ref{fig:attitude_compare}}) execute the maneuver with identical kinematic pacing, verifying that the differentiable environment matches the integration fidelity of classical successive convexification without numerical drift.

\begin{figure}[htbp]
    \centering
    \includegraphics[width=0.8\textwidth]{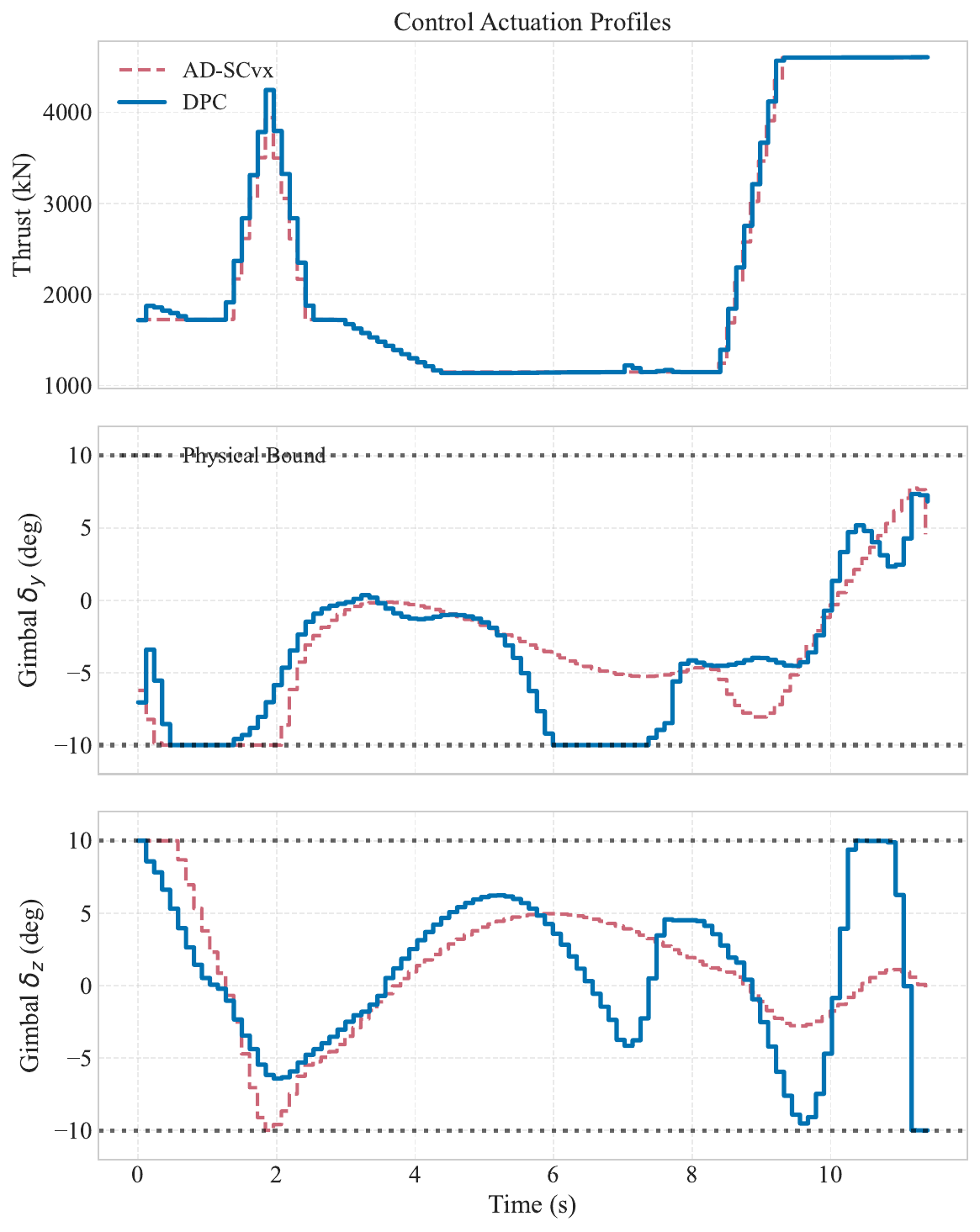}
    \caption{Control actuator responses for the nominal scenario. DPTC exhibits more pronounced ``bang-bang'' behavior in the pitch gimbal angle ($\delta_y$) compared to the slightly smoothed SCvx solution.}
    \label{fig:gimbal_thrust}
\end{figure}

\textbf{2. Actuation Allocation and Fuel Optimality} \\
Fuel-optimal landing under bounded control authority typically yields bang-bang actuation (\textbf{Fig.~\ref{fig:gimbal_thrust}}). Both methods produce virtually identical thrust profiles. However, slight divergences appear in the gimbal angle allocations ($\delta_y, \delta_z$). DPTC ($N_p=1$) exhibits more pronounced bang-bang behavior, frequently saturating the $\pm 10^\circ$ limits, while SCvx shows slightly smoother transitions. This discrepancy arises because SCvx employs interior-point barrier functions that induce constraint-repulsion smoothing, whereas DPTC evaluates exact projection operators ($\text{Proj}_{\mathcal{U}}$) without artificial barriers, operating directly on the absolute boundaries.

\begin{figure}[htbp]
    \centering
    \includegraphics[width=0.6\textwidth]{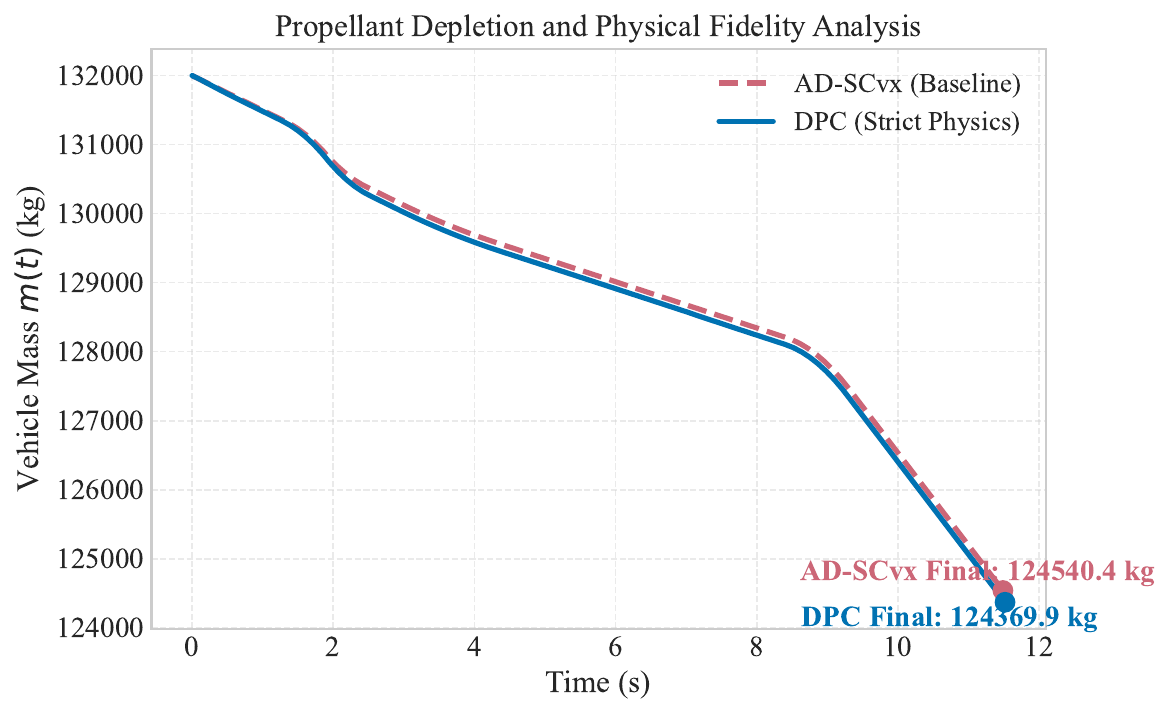}
    \caption{Vehicle mass depletion over the nominal trajectory, showing a negligible difference of less than $0.25\%$.}
    \label{fig:nominal_mass}
\end{figure}

\textbf{3. Mass Depletion and Limitations of Deterministic Optimization} \\
Both strategies achieve equivalent fuel efficiency (\textbf{Fig.~\ref{fig:nominal_mass}}), with final vehicle masses of $124,556.9$ kg (CS-AD-SCvx) and $124,238.6$ kg (DPTC). This negligible difference ($<0.25\%$) confirms convergence to similar fuel-optimal performance levels.

However, this deterministic analysis highlights a practical operational limitation. While boundary-active bang-bang control minimizes fuel consumption, it exhausts control authority. Under closed-loop conditions, exogenous gusts could easily precipitate actuator saturation and attitude divergence. This limitation highlights the necessity of unlocking the full Lagrangian ensemble ($N_p > 1$) in subsequent sections, demonstrating how DPTC utilizes distribution shaping to preserve control margins under stochastic disturbances.
\subsection{Robust Distribution Shaping under Aerodynamic Uncertainties}
\label{subsec:robust_planning}

Having established the deterministic baseline, this section utilizes the full Lagrangian particle ensemble ($N_p=32$) to evaluate the robustness of the proposed frameworks. A $5\%$ unsteady aerodynamic disturbance ($\boldsymbol{w}$) is injected into the nonlinear flip maneuver (see \ref{app:hyperparameters}. Rather than analyzing singular trajectories, this section investigates the spatiotemporal evolution of the $3\sigma$ dispersion tubes (capturing 99.7\% of the off-nominal dispersions) generated by the CS-AD-SCvx baseline and the proposed DPTC.

\begin{figure}[htbp]
    \centering
    \includegraphics[width=0.9\textwidth]{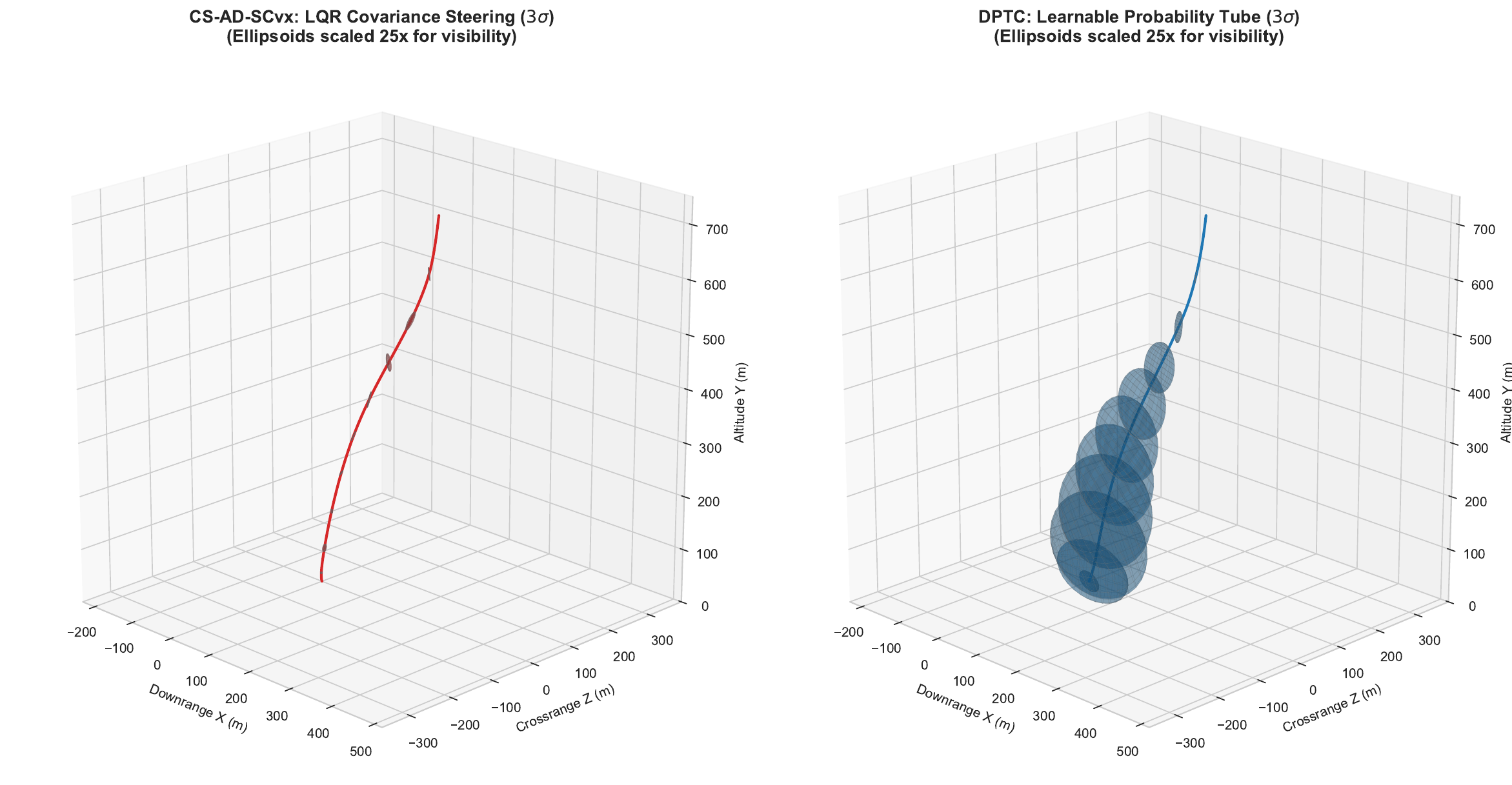}
    \caption{Comparative 3D spatial trajectories and corresponding $3\sigma$ covariance ellipsoids under $5\%$ aerodynamic disturbances. The ellipsoids are scaled by a factor of 25x to render the topological differences visible across the flight domain.}
    \label{fig:3d_tubes}
\end{figure}

\begin{figure}[htbp]
    \centering
    \includegraphics[width=0.7\textwidth]{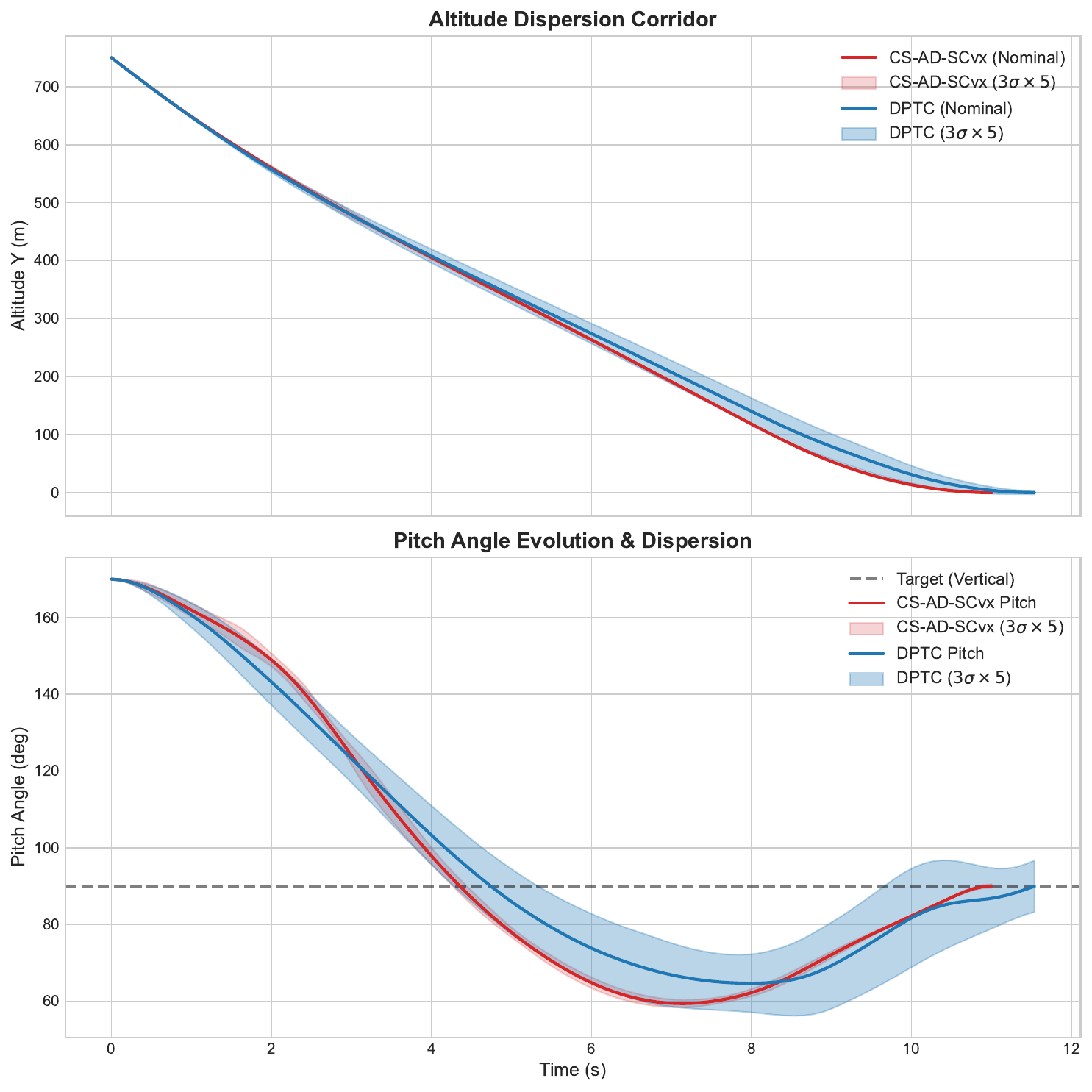}
    \caption{State dispersion corridors for altitude (top) and pitch angle (bottom). The $3\sigma$ boundaries are visually magnified by $10\times$ to highlight the structural differences between rigid state suppression (CS-AD-SCvx) and adaptive distribution shaping (DPTC).}
    \label{fig:state_corridors}
\end{figure}

\textbf{1. Spatial Dispersion and Distribution Shaping} \\
The handling of non-Gaussian uncertainties differs significantly between the two methodologies, which is reflected in their spatial dispersion tubes. \textbf{Fig.~\ref{fig:3d_tubes}} visualizes the 3D spatial covariance ellipsoids, and \textbf{Fig.~\ref{fig:state_corridors}} details the altitude and pitch angle corridors. 

To ensure visibility against the macroscopic $750$-meter altitude sweep, the $3\sigma$ covariance boundaries are visually scaled ($25\times$ for 3D ellipsoids and $5\times$ for 2D corridors). This scaling highlights the structural differences in how each algorithm propagates state dispersions.

As observed, the spatial tube generated by CS-AD-SCvx is tightly constrained, whereas DPTC exhibits a noticeably wider spatial dispersion envelope, particularly in the mid-flight altitude corridor (\textbf{Fig.~\ref{fig:state_corridors}}). 

This structural divergence illustrates the limitations of the local Gaussian approximation inherent in classic Certainty Equivalence. Synthesized under the unconstrained Separation Principle, the Riccati-based feedback strictly minimizes local state deviations under the assumption of infinite control authority. And, this yields a tightly constrained spatial dispersion tube that is mathematically optimal in the unconstrained domain, yet physically infeasible when subjected to strict actuator limits.

In contrast, DPTC employs a constraint-aware distribution shaping strategy. Driven by the continuous terminal loss landscape and backpropagated saturation penalties, DPTC prioritizes control effort allocation over rigid path tracking. During the high-dynamic-pressure pitch-up maneuver, DPTC relaxes spatial tracking requirements, permitting the dispersion manifold to expand within safe boundaries. This spatial relaxation conserves critical control margins, which are subsequently deployed during the terminal descent phase to concentrate the state distribution and satisfy landing precision requirements.

\begin{figure}[htbp]
    \centering
    \includegraphics[width=0.8\textwidth]{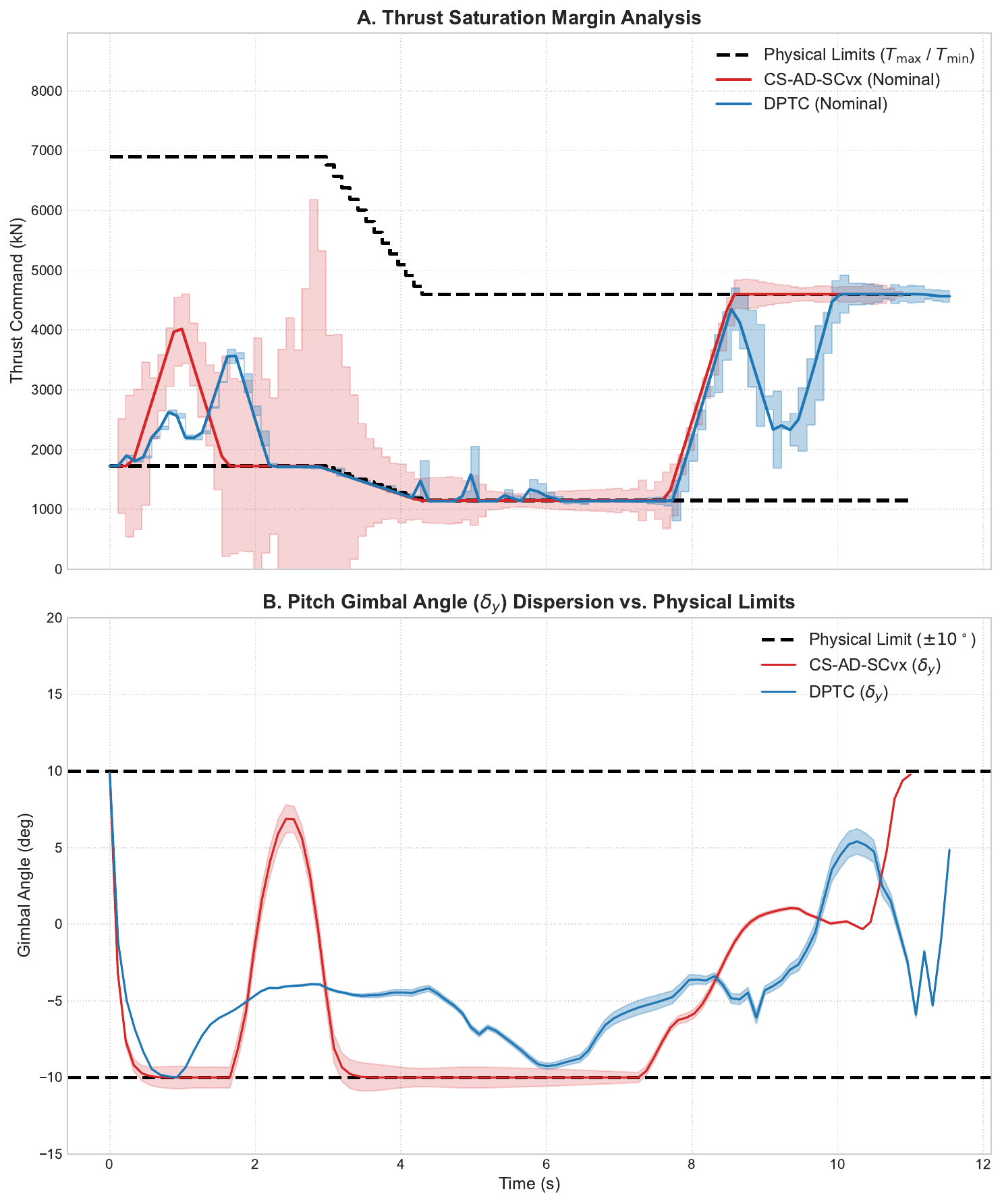}
    \caption{Actuator Saturation Margin Analysis. The shaded regions denote the $3\sigma$ unconstrained control demand. Driven by unconstrained Riccati feedback, CS-AD-SCvx exhibits unconstrained demands that exceed physical limits. Conversely, DPTC actively satisfies boundary constraints.}
    \label{fig:actuator_saturation}
\end{figure}

\textbf{2. Actuator Saturation and Constraint-Awareness} \\
The operational implications of this spatial relaxation become evident when mapping the $3\sigma$ dispersion tubes into the actuator command space. \textbf{Fig.~\ref{fig:actuator_saturation}} illustrates the control demand envelopes. It is crucial to clarify that the shaded regions represent the unconstrained mathematical control demand ($\delta \boldsymbol{u} = -\boldsymbol{K} \delta \boldsymbol{x}$) calculated via linear feedback projection, not the physically executed actuation.

\textit{Thrust Deep-Throttling Limit ($T_{\min}$):} As shown in \textbf{Fig.~\ref{fig:actuator_saturation}A}, CS-AD-SCvx violates the deep-throttling limit between $t = 1.0$ s and $4.0$ s. To enforce the tightly constrained spatial tube, the unconstrained LQR commands the engine below $T_{\min}$. Because the theoretical covariance propagation does not account for hardware truncation, this unconstrained demand renders the closed-loop tracking physically infeasible, invalidating the theoretical covariance tube.

The $3\sigma$ unconstrained demand of DPTC exhibits minor, transient boundary violations near the terminal phase. However, this reflects the algorithm's constraint-aware formulation. Because the truncation operators ($\text{Proj}_{\mathcal{U}}$) are embedded directly into the computational graph during offline training, the optimizer accounts for these saturation events. The framework permits the pre-truncation demand of tail-particles to momentarily exceed limits, as the resulting physical clamping does not destabilize the globally optimized distribution. By exploiting the relaxed spatial corridor, DPTC modulates the thrust profile to operate safely along the physical boundaries.

\textit{Gimbal Angle Authority:} The attitude control analysis (\textbf{Fig.~\ref{fig:actuator_saturation}B}) highlights an additional vulnerability in the unconstrained baseline. Between $t = 1.5$ s and $6.0$ s, the CS-AD-SCvx $3\sigma$ gimbal demand operates at the $-10^\circ$ physical limit. Under these conditions, subsequent exogenous disturbances would fully exhaust Thrust Vector Control (TVC) authority, potentially leading to attitude divergence. 

Conversely, DPTC implicitly suppresses the feedback gains ($\boldsymbol{K}$) during the high-dynamic-pressure phase via backpropagation, maintaining the $3\sigma$ gimbal dispersion strictly within the $\pm 10^\circ$ physical limits. 

These evaluations indicate that DPTC achieves a practical engineering trade-off. By relaxing spatial tracking constraints, the framework prevents actuator saturation, ensuring control authority and vehicle stability under bounded stochastic disturbances.

\subsection{Closed-Loop Monte Carlo Validation: Landing Dispersion and Survivability}
\label{subsec:landing_footprint}

To evaluate the disturbance-rejection capabilities of both offline-optimized guidance policies against unmodeled aerodynamic uncertainties, this section extracts the landing footprints from $N=5000$ closed-loop Monte Carlo (MC) flight simulations. In these runs, the vehicle performs trajectory tracking relying exclusively on the pre-computed nominal trajectories ($\boldsymbol{X}_{\text{ref}}$, $\boldsymbol{U}_{\text{ref}}$) and the time-varying affine feedback gain matrices ($\boldsymbol{K}$). The online feedback commands ($\delta \boldsymbol{u} = -\boldsymbol{K} \delta \boldsymbol{x}$) are strictly subjected to the physical actuator hard constraints ($T \in [T_{\min}, T_{\max}]$, $\delta \in [\pm 10^\circ]$) without any online re-planning. 

\begin{figure}[htbp]
    \centering
    \begin{minipage}{0.48\textwidth}
        \centering
        \includegraphics[width=\linewidth]{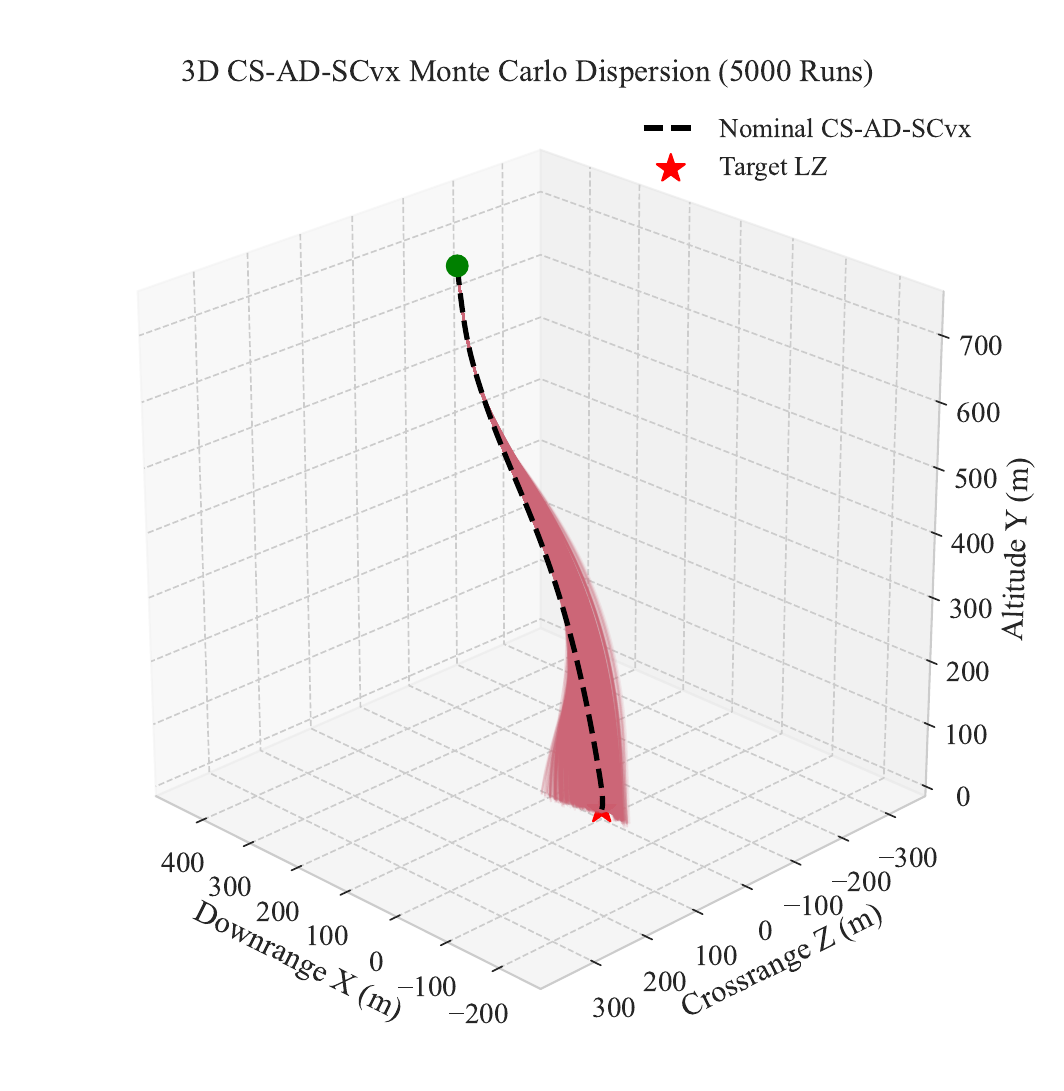}
    \end{minipage}\hfill
    \begin{minipage}{0.48\textwidth}
        \centering
        \includegraphics[width=\linewidth]{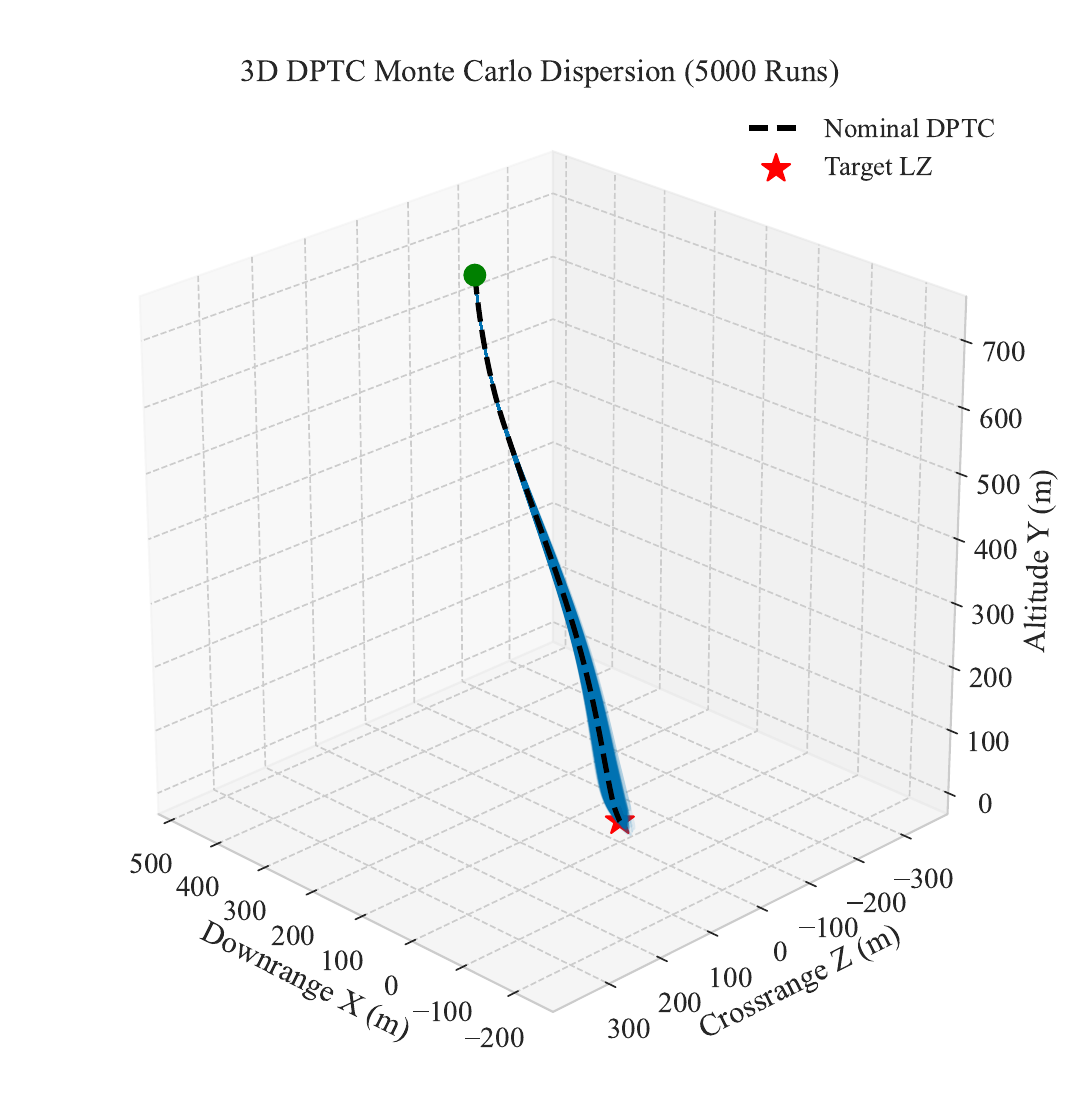}
    \end{minipage}
    \caption{Closed-loop Monte Carlo 3D trajectories ($N=5000$) under $5\%$ aerodynamic uncertainties. The CS-AD-SCvx exhibits catastrophic divergence due to actuator saturation, whereas the DPTC maintains a tightly bundled descent profile.}
    \label{fig:mc_spaghetti}
\end{figure}

\begin{figure}[htbp]
    \centering
    \begin{minipage}{0.48\textwidth}
        \centering
        \includegraphics[width=\linewidth]{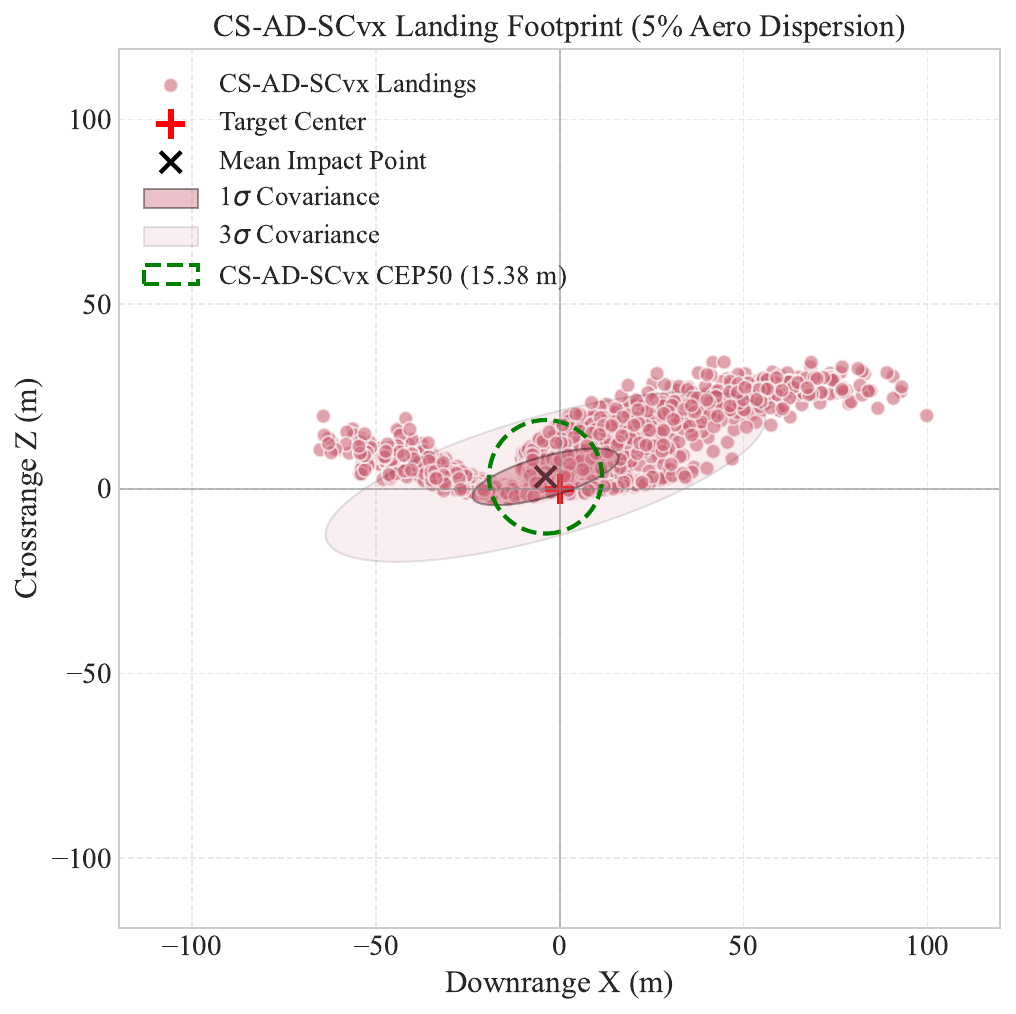}
    \end{minipage}\hfill
    \begin{minipage}{0.48\textwidth}
        \centering
        \includegraphics[width=\linewidth]{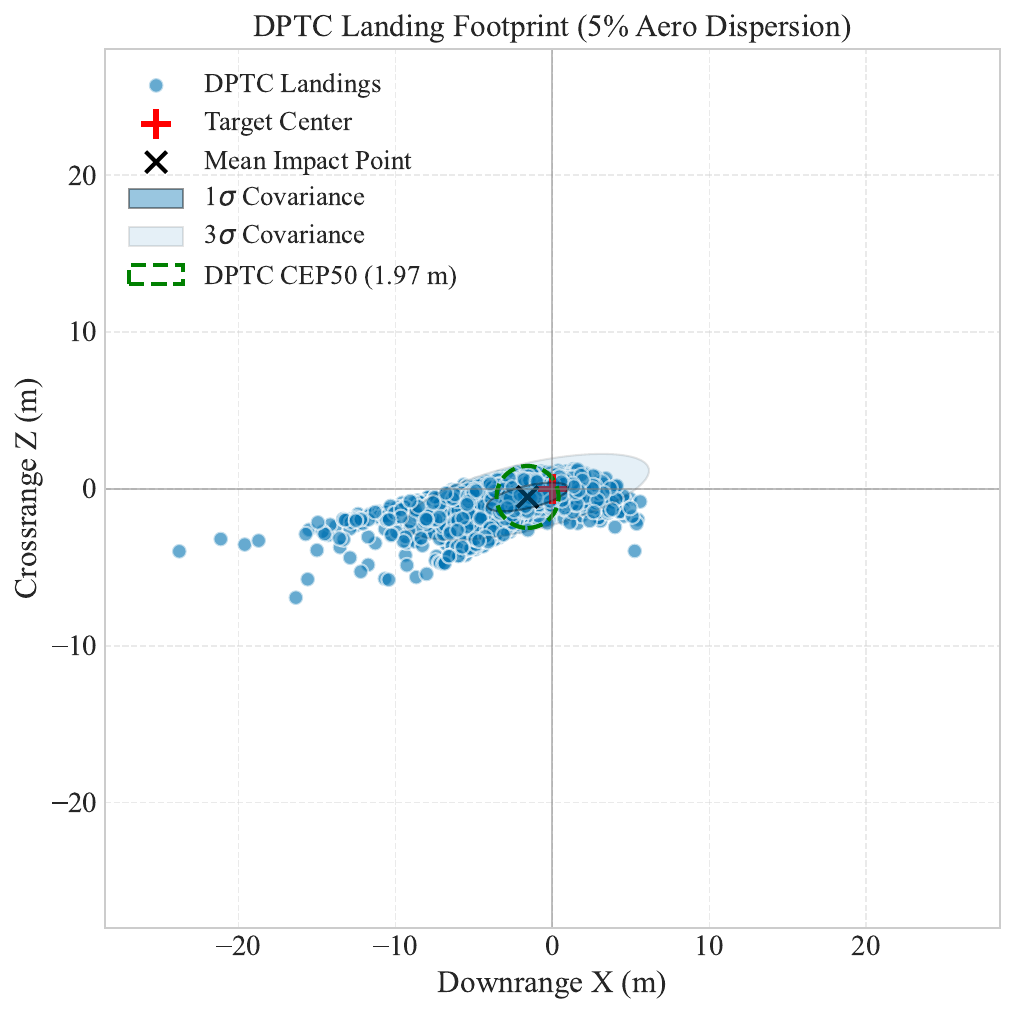}
    \end{minipage}
    \caption{Terminal landing footprint dispersion in the horizontal X-Z plane. The DPTC reduces the CEP$_{50}$ by $87\%$ and exhibits a highly structured covariance ellipse, reflecting its proactive energy management and cross-coupling perception capabilities.}
    \label{fig:mc_footprint}
\end{figure}

The macroscopic flight robustness is visualized via the 3D trajectory dispersions (\textbf{Fig.~\ref{fig:mc_spaghetti}}), while the terminal precision is quantified by the landing footprints and the 50\% Circular Error Probable (CEP$_{50}$) in the X-Z plane (\textbf{Fig.~\ref{fig:mc_footprint}}). The results demonstrate that the dispersion topologies of the two methods exhibit distinct physical characteristics under severe control saturation.

\textbf{1. Asymmetric Dispersion under Unconstrained Feedback (CS-AD-SCvx):} \\
As shown in the 3D dispersion plots (\textbf{Fig.~\ref{fig:mc_spaghetti}}, Left), the CS-AD-SCvx baseline guided by the Time-Varying LQR exhibits significant spatial dispersion during the high-dynamic-pressure phase. A distinct topological phenomenon is observed: the Monte Carlo dispersion cloud systematically drifts to one side of the nominal trajectory, resulting in an asymmetric envelopment. The terminal footprint (\textbf{Fig.~\ref{fig:mc_footprint}}, Left) manifests as a broad and off-centered dispersion cloud, yielding a CEP$_{50}$ of $15.38$ m. 

This biased dispersion originates from the asymmetric effects of actuator saturation on an unconstrained feedback policy. Because the TVLQR gain is synthesized under the assumption of unconstrained control authority, the resulting feedback policy frequently utilizes a substantial portion of the available actuator authority, leaving only limited and asymmetric control margins near the operational boundaries. When aerodynamic disturbances require corrective actions toward a saturated direction, the commanded inputs are clipped by the actuator limits, creating a mismatch between the assumed linear covariance evolution and the physically executed closed-loop dynamics. This mismatch is subsequently amplified through nonlinear cross-axis coupling, ultimately leading to the asymmetric landing dispersion observed in the Monte Carlo simulations.

\textbf{2. Symmetric Envelopment and Distribution Shaping (DPTC):} \\
In contrast, the DPTC algorithm demonstrates stable convergence under the same disturbances (\textbf{Fig.~\ref{fig:mc_spaghetti}}, Right). Topologically, the off-nominal trajectories symmetrically envelop the nominal baseline, forming a concentrated spatial tube. The landing points are densely clustered, yielding a $87\%$ reduction in CEP$_{50}$ ($1.97$ m). The dispersion morphology manifests as a structured, flattened ellipse (\textbf{Fig.~\ref{fig:mc_footprint}}, Right), reflecting the physical characteristics of the saturation-aware distribution shaping policy.

The symmetric envelopment and the elongated terminal covariance ellipse ($\in [-15, 5]$ m downrange, $\in [-5, 2]$ m crossrange, with a $15^\circ$ inclination) highlight the efficacy of the DPTC policy in two operational aspects:
\begin{itemize}
    \item \textbf{Energy Management and Controlled Degradation:} In the deep-throttle saturation regime near touchdown, correcting downrange (X-axis) deviations necessitates pitching the vehicle, which diverts the vertical thrust component required to oppose gravity. Because the sub-differentiable clamping operators are embedded during offline training, the DPTC policy accounts for this actuation penalty. Under extreme adverse wind shear, DPTC executes a controlled performance trade-off: it intentionally sacrifices marginal horizontal precision to prioritize the terminal vertical soft-landing attitude and velocity constraints, thereby preventing hard landing violations.
    
    \item \textbf{Aerodynamic Cross-Coupling Perception:} The $15^\circ$ inclination of the covariance ellipse indicates that the backpropagation-derived gain matrix $\boldsymbol{K}$ captures the aerodynamic cross-coupling derivatives (e.g., sideslip-to-pitch dynamic coupling). It dynamically allocates damping effort in alignment with the natural manifold structure of the unsteady flow field, rather than treating orthogonal Cartesian axes independently as in the traditional unconstrained LQR.
\end{itemize}

In summary, by incorporating physical clamping functions and exact nonlinear ODE integration into the optimization computational graph, DPTC mitigates the limitations of unconstrained closed-loop guidance near operational boundaries. This results in significantly tighter landing dispersions and enhanced physical constraint satisfaction under severe non-Gaussian aerodynamic disturbances.

\subsection{Computational Complexity and Algorithmic Scalability}
\label{subsec:computational_complexity}

The computational efficiency and algorithmic scaling behaviors of the tested methodologies were evaluated on a standard hardware configuration (Intel Core i7-13620H CPU, NVIDIA RTX 4060 GPU).

\textbf{1. Complexity of Successive Convexification} \\
The conventional SCvx architecture relies on sequential execution and interior-point solvers. For the nominal flip-landing maneuver, the deterministic AD-SCvx baseline required 46 iterations ($79.01$ seconds) to converge. Incorporating robust covariance steering (CS-AD-SCvx) increased the requirement to 49 iterations and $106.56$ seconds. This $\approx 35\%$ computational overhead is driven by the analytical extraction of state and control Jacobians ($\boldsymbol{A}_k, \boldsymbol{B}_k$) via automatic differentiation and the backward integration of the Riccati equations. Furthermore, the sequential nature of SOCP solvers fundamentally limits the scalability of this approach for large-scale probability evaluations.

\textbf{2. DPTC Scalability and Hardware Utilization} \\
In contrast, the DPTC framework evaluates the un-linearized dynamics using parallel tensor operations. For an ensemble of $N_p = 32$ particles, a 50-epoch optimization required $122.01$ seconds on the CPU, making it competitive with a single CS-AD-SCvx run for moderate batch sizes. 

Transitioning to a GPU architecture introduces significant scaling benefits for larger ensembles. Executing 50 epochs with $N_p = 32$ particles required $343.24$ seconds on the GPU, while scaling the ensemble size to $N_p = 1024$ particles required $342.12$ seconds. This $\mathcal{O}(1)$ scaling behavior with respect to batch size is achieved via GPU vectorization (\texttt{torch.vmap}), which evaluates the ensemble rollouts in parallel. The CPU outperforms the GPU for smaller batches because the 100-step Runge-Kutta integration is inherently sequential; communication latency across the PCIe bus dominates the computational time for small state dimensions on the GPU. Conversely, modern CPUs execute these sequential loops efficiently within their local cache.

\textbf{3. Offline Training vs. Online Deployment} \\
Although DPTC requires a higher total offline computational cost (typically thousands of epochs) to reach convergence compared to a single SCvx run, it shifts the computational burden entirely to the offline training phase. Once the unified optimization graph converges, the synthesized joint control policy $(\boldsymbol{U}_{\text{ref}}, \mathcal{K})$ is explicitly stored as a time-varying tensor sequence. That means the online deployment reduces to an $\mathcal{O}(1)$ lookup and basic matrix multiplication, making the resulting constraint-aware policy highly suitable for high-frequency, real-time onboard execution.
\section{Conclusions}
\label{sec:conclusions}

This paper investigated the high-angle-of-attack flip maneuver of reusable launch vehicles, establishing a unified differentiable physics framework for robust trajectory optimization under aerodynamic uncertainties. By integrating a neural aerodynamic surrogate into a differentiable 6-DoF flight simulation environment, the framework enables exact gradient propagation through nonlinear dynamics. Using this architecture, two distinct mathematical paradigms for robust guidance were evaluated: local differentiable convexification (CS-AD-SCvx) and Differentiable Particle Tube Control (DPTC). The main conclusions are summarized as follows:

\textbf{1. Limitations of Unconstrained Covariance Steering under Actuator Saturation (CS-AD-SCvx):} \\
The CS-AD-SCvx baseline utilizes automatic differentiation to extract highly accurate local gradients for successive convex optimization. Under nominal conditions, it achieves theoretical fuel optimality. However, closed-loop Monte Carlo simulations reveal the operational limitations of linear covariance steering under severe aerodynamic dispersions. Synthesized under the Separation Principle, the unconstrained covariance tubes do not account for rigid physical actuator limits during feedback design. When the required corrective feedback exceeds the deep-throttling or gimbal limits, the physically executed commands are truncated. This truncation invalidates the theoretical closed-loop covariance propagation, demonstrating that local linear uncertainty propagation is highly sensitive at the boundaries of the operational envelope.

\textbf{2. Robust Guidance via Differentiable Distribution Shaping (DPTC):} \\
To address the limitations of separated trajectory generation and unconstrained tracking, the proposed DPTC framework reformulates robust guidance as a constrained stochastic distribution shaping problem. Instead of relying on local Gaussian covariance assumptions, DPTC represents the evolving uncertainty distribution through a Lagrangian particle ensemble, optimizing the closed-loop probability manifold under the exact nonlinear dynamics. 

By embedding sub-differentiable clamping operators (actuator saturation) directly into the computational graph, DPTC jointly synthesizes the nominal feedforward sequence and the time-varying affine feedback policy. Monte Carlo validations demonstrate that this saturation-aware policy actively executes a controlled performance trade-off---sacrificing marginal horizontal precision to strictly satisfy terminal vertical velocity and attitude constraints. This constraint-aware formulation prevents boundary violations and reduces the landing footprint CEP$_{50}$ by $87\%$ under severe wind shear.

\textbf{3. Scalable Optimization Architecture for Advanced GNC Systems:} \\
Beyond the specific application of reusable rocket recovery, the proposed differentiable framework offers a scalable and rigorous optimization architecture for high-dimensional, nonlinear aerospace guidance problems:

\begin{itemize}
    \item \textit{End-to-End Sensitivity Propagation:} Unlike traditional sampling-based stochastic programming or finite-difference numerical methods, the framework exploits a natively differentiable physical environment to compute exact analytical gradients across complex aerodynamic transients. This allows the feedback policy to be optimized directly via physical sensitivities, providing a stable, gradient-driven pathway for safety-critical aerospace systems.
    
    \item \textit{Physics-Constrained Ensemble Control:} Rather than relying on idealized linear covariance propagation, the framework manages non-Gaussian uncertainty by directly shaping the macroscopic statistical evolution of a particle ensemble. This probability evolution is strictly governed by actual rigid-body kinematics, unsteady aerodynamic forces, and physical actuator bounds, rather than artificial mathematical approximations.
\end{itemize}

In conclusion, this work presents a transition from deterministic trajectory tracking to physics-informed stochastic distribution shaping. The proposed framework provides an alternative and highly practical formulation for advanced aerospace Guidance, Navigation, and Control (GNC) systems, demonstrating that integrating differentiable programming directly with exact flight mechanics can significantly enhance mission robustness and physical constraint satisfaction in highly constrained flight environments.

\section*{Acknowledgments}
This work was supported by the National Natural Science Foundation of China under Grant No. 92470120.

\bibliographystyle{elsarticle-num}
\bibliography{refs}

\appendix
\section{Differentiable Aerodynamic Surrogate: CFD Dataset and MLP Architecture}
\label{app:aerodynamic_surrogate}

The high computational cost of traditional Reynolds-Averaged Navier-Stokes (RANS) simulations poses a prohibitive barrier for optimal control tasks that require tight, high-frequency coupling between fluid flow and rigid-body dynamics. Furthermore, classical tabular interpolation schemes (e.g., look-up tables) lack the analytical smoothness necessary for Jacobian extraction and continuous gradient backpropagation. To overcome these limitations, a fully differentiable Multi-Layer Perceptron (MLP) surrogate was developed, rooted in a comprehensive high-fidelity CFD dataset.

\subsection{CFD Data Acquisition and Grid Topology}
The aerodynamic dataset spans the extreme operational envelope of the Starship S31 prototype during its characteristic flip-and-landing maneuver. The vehicle configuration was fixed with the forward flaps deployed flat and the aft flaps folded upward at an $80^\circ$ angle. Because the Mach number remains strictly below $0.3$ during the terminal descent and pitch-up phase, the flow was modeled as incompressible. 

\begin{figure}[htbp]
    \centering
    \includegraphics[width=0.8\textwidth]{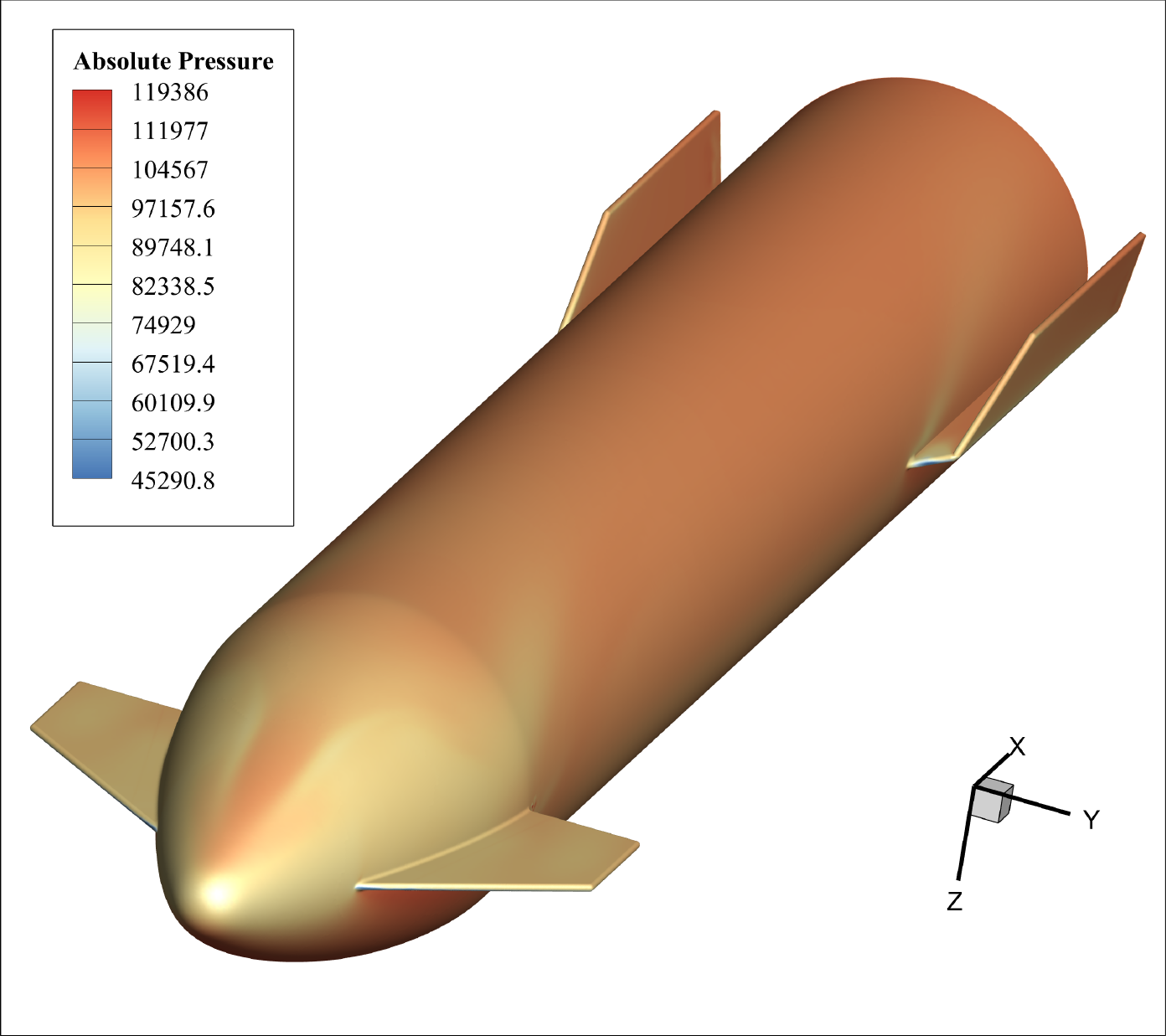}
    \caption{Computational grid topology (boundary-layer resolved) and the resulting surface pressure distribution of the Starship S31 prototype under an extreme belly-flop angle of attack ($\alpha = 50^\circ$).}
    \label{fig:cfd_contour}
\end{figure}

Steady RANS simulations were executed utilizing the open-source SU2 solver. The computational domain was discretized using boundary-layer-resolved unstructured grids comprising approximately $4.0 \times 10^6$ cells. To capture the full rotational envelope of the flip maneuver, the aerodynamic force and moment coefficients were sampled across the complete $360^\circ$ spectrum (from $0^\circ$ to $350^\circ$) at intervals of $10^\circ$. \textbf{Fig.~\ref{fig:cfd_contour}} visually demonstrates the grid resolution and the highly nonlinear surface pressure distribution induced by massive flow separation at $\alpha = 50^\circ$.
\subsection{MLP Architecture and Training Methodology}
The discrete CFD dataset was subsequently utilized to train a compact yet highly expressive MLP surrogate. To strictly enforce spatial periodicity and avoid the discontinuities inherent in angular wrap-around (at $0^\circ$ and $360^\circ$), the raw Angle of Attack (AoA, $\alpha$) was feature-engineered into its trigonometric basis $[\cos\alpha, \sin\alpha]^T$. 

The neural architecture and hyperparameter selections are detailed in \textbf{Table~\ref{tab:mlp_architecture}}. Prior to training, the target aerodynamic coefficients ($C_l, C_d, C_m$) were standardized using zero-mean and unit-variance scaling to accelerate gradient convergence. The dataset was partitioned with an $80/20$ train-test split, and the network weights were optimized using the Adam algorithm governed by a Mean Squared Error (MSE) loss criterion, augmented with an $L_2$ weight decay regularization ($\lambda = 10^{-5}$) to further penalize high-frequency overfitting.

\begin{table}[htbp]
    \centering
    \caption{MLP Surrogate Architecture and Training Configurations}
    \label{tab:mlp_architecture}
    \begin{tabular}{ll}
        \toprule
        \textbf{Parameter} & \textbf{Specification} \\
        \midrule
        Input Features & $[\cos\alpha, \sin\alpha]^T \in \mathbb{R}^2$ \\
        Output Features & $[C_l, C_d, C_m]^T \in \mathbb{R}^3$ \\
        Hidden Layers & Two dense layers [64, 64] neurons \\
        Activation Function & Shifted Softplus ($f(x) = \ln(1+e^x) - \ln2$) \\
        Optimizer & Adam ($lr = 1 \times 10^{-3}$, weight decay $= 10^{-5}$) \\
        Loss Function & Mean Squared Error (MSE) \\
        Data Normalization & Standard Scaler ($\mu = 0, \sigma^2 = 1$) \\
        Epochs & 500 \\
        \bottomrule
    \end{tabular}
\end{table}

\subsection{Algorithmic Significance: Smoothness, Monotonicity, and Gradient Propagation}
While the MLP exhibits excellent regression fidelity relative to the original CFD discrete samples, its most critical contribution to the proposed optimization frameworks (both CS-AD-SCvx and DPTC) lies in its \textit{differentiable smoothing properties}. 

\begin{figure}[htbp]
    \centering
    \includegraphics[width=0.9\textwidth]{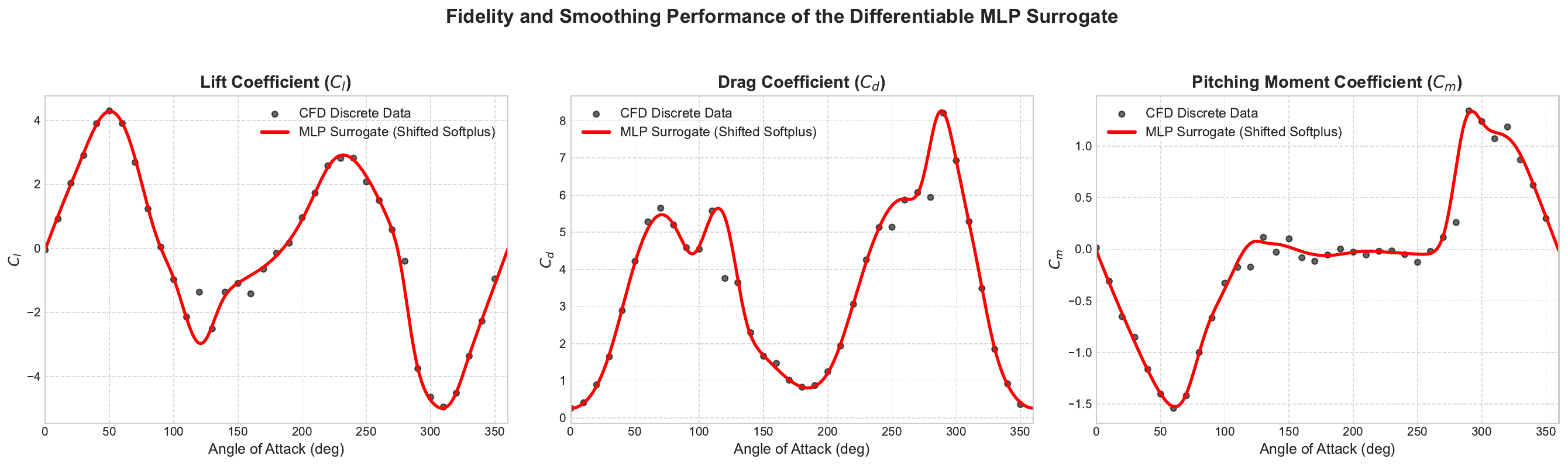}
    \caption{Fidelity and smoothing performance of the MLP surrogate across the full $360^\circ$ AoA spectrum. The neural surrogate (solid lines) successfully captures the global nonlinear dynamics of the discrete CFD samples (scatter points) while inherently filtering out high-frequency numerical noise.}
    \label{fig:mlp_fitting}
\end{figure}

As illustrated in \textbf{Fig.~\ref{fig:mlp_fitting}}, numerical noise and localized non-physical oscillations—often inevitable in steady RANS solutions for massively separated bluff-body flows—are present in the raw discrete dataset. Rather than overfitting these high-frequency anomalies, the MLP acts as a nonlinear low-pass filter, projecting the noisy data onto a globally smooth $\boldsymbol{C^\infty}$ continuous manifold. 

The achievement of this ideal manifold heavily relies on the mathematically deliberate choice of the \textit{Shifted Softplus} activation function ($f(x) = \ln(1+e^x) - \ln 2$), which specifically addresses the critical flaws of standard alternatives in automatic differentiation contexts:
\begin{itemize}
    \item \textbf{Overcoming $C^1$-Discontinuity (vs. ReLU):} While the ubiquitous Rectified Linear Unit (ReLU) is standard in static pattern recognition, it yields a piecewise linear manifold that is $C^1$-discontinuous at the origin. When composed through hundreds of Runge-Kutta integration steps, a frozen ReLU network generates an exponentially rugged loss landscape populated with gradient kinks. The Shifted Softplus provides infinite differentiability ($C^\infty$), eliminating these kinks.
    \item \textbf{Enforcing Physical Plausibility (vs. SiLU/GELU):} To achieve $C^\infty$ smoothness, modern variants like SiLU or GELU are frequently employed. However, these functions exhibit localized non-monotonic behavior for negative inputs (i.e., a subtle negative dip). In aerodynamic surrogate modeling, this non-monotonicity risks introducing artificial, non-physical oscillatory artifacts (wiggles) into the force coefficients. The Shifted Softplus is strictly monotonically increasing ($f'(x) > 0$), thereby guaranteeing absolute physical rigidity while maintaining a zero-centered origin ($f(0)=0$) to preserve optimal initialization variance.
\end{itemize}

This mathematically rigorous smoothing capability is algorithmically paramount. For the baseline CS-AD-SCvx methodology, it strictly prevents the extracted aerodynamic Jacobians ($\boldsymbol{A}_k, \boldsymbol{B}_k$) from exhibiting chaotic, discontinuous step-jumps that would otherwise violently derail the interior-point solver's trust-region mechanism. For the proposed DPTC framework, this smooth, strictly monotonic aerodynamic landscape provides a singularity-free loss manifold, ensuring stable and robust Backpropagation Through Time (BPTT) for the probability transport engine.

\subsection{Algorithmic Reproducibility and Hyperparameter Configuration}
\label{app:hyperparameters}

To ensure the full reproducibility of the comparative trajectory optimization results and to clarify the exact formulations underlying the computational graphs, the critical hyperparameters and implementation specifics are detailed as follows:

\textbf{1. Aerodynamic Disturbance Model} \\
The 5\% unsteady aerodynamic uncertainty evaluated in the Monte Carlo simulations is modeled as a multiplicative Gaussian noise applied directly to the neural aerodynamic predictions. The exact injection takes the form of $C_{(\cdot)} = C_{(\cdot),\text{pred}} \times (1 + 0.05 w)$, where $w \sim \mathcal{N}(0, 1)$ represents the standard normal distribution sampled independently for the lift, drag, and pitch moment coefficients at each integration step.

\textbf{2. Differentiable Optimization Setup (Proposed Framework)} \\
The Differentiable Probability Transport Control (DPTC) computational graph is optimized using the Adam algorithm. The optimization initializes with a learning rate of $5 \times 10^{-3}$ and employs a \texttt{ReduceLROnPlateau} scheduler (decay factor = $0.5$, patience = $150$). The optimization is executed for $5000$ epochs. For the unconstrained loss functional ($\mathcal L_{\mathrm{total}}$), the feedback regularization weight is set to $\gamma = 1.0$. The tail-risk surrogate adopts the exact continuous form $\phi(z) = \mathrm{ReLU}^2(z)$. To ensure absolute survivability without suffering from gradient vanishing, the tail-risk weight $\lambda_r$ is tiered: hard physical bounds (e.g., maximum/minimum engine thrust magnitude, gimbal actuator limits) are heavily penalized with $\lambda_r = 10^5$, whereas operational constraints (e.g., aerodynamic pitch corridor, engine throttling rate) are scaled by $\lambda_r = 10^4$.

\textbf{3. CS-AD-SCvx Baseline Configuration} \\
For the successive convexification baseline utilizing local covariance steering, the probabilistic chance constraints are strictly bounded by a $3\sigma$ confidence margin ($\kappa = 3.0$). This theoretical margin targets an approximate $99.7\%$ constraint satisfaction probability under the local Gaussian linear propagation assumption. The trust-region radius is dynamically updated via an annealing schedule with a strict lower bound to prevent premature numerical deadlock during the highly nonlinear flip maneuver.

\end{document}